\documentclass[11pt]{article}
\usepackage{arxiv}

\usepackage[T1]{fontenc}
\usepackage[utf8]{inputenc}
\usepackage{amsmath,amssymb}
\usepackage{booktabs}
\usepackage{float}
\usepackage{array}
\usepackage{multirow}
\usepackage{graphicx}
\usepackage{xcolor}
\usepackage{tikz}
\usetikzlibrary{positioning,arrows.meta,calc,decorations.pathmorphing,fit}
\usepackage[most]{tcolorbox}
\usepackage{soul}
\usepackage{url}
\usepackage[hidelinks]{hyperref}

\definecolor{newbg}{RGB}{212,235,250}
\newenvironment{nw}{}{}
\newenvironment{nwf}{}{}
\sethlcolor{newbg}
\soulregister\ref7
\soulregister\cite7
\soulregister\textit7
\newcommand{\hlnew}[1]{#1}

\newcommand{\orcidlink}[1]{\href{https://orcid.org/#1}{\textsuperscript{\textsf{iD}}\,\texttt{#1}}}

\hypersetup{
  pdfauthor={Ba Tu Phung and Huy Hoang Le and Kim-Anh Nguyen},
  pdftitle={Understanding the limits of snapshot-based fault diagnosis in distribution networks},
  pdfkeywords={fault diagnosis, distribution network reliability, snapshot measurements, machine learning, high-impedance faults, measurement sparsity, sensor placement, phasor measurement noise},
  pdfsubject={Electrical Engineering and Systems Science - Systems and Control (eess.SY)},
  pdfcreator={LaTeX with arxiv.sty},
}

\begin{document}

\title{Understanding the limits of snapshot-based fault diagnosis in distribution networks}

\author{%
\textbf{Ba Tu Phung}\textsuperscript{1}\,\orcidlink{0009-0000-1126-8334}\\
\textbf{Huy Hoang Le}\textsuperscript{2}\,\orcidlink{0009-0009-2138-5468}\\
\textbf{Kim-Anh Nguyen}\textsuperscript{1,*}\,\orcidlink{0000-0003-3408-847X}\\[7pt]
\normalfont
\small\textsuperscript{1}Faculty of Electrical Engineering, University of Science and Technology,\\
\small The University of Danang, 54 Nguyen Luong Bang Street, Da Nang 550000, Vietnam\\
\small\textsuperscript{2}R\&D Department, PowerMore Ltd., Da Nang 550000, Vietnam\\
\small\textsuperscript{*}Corresponding author: \href{mailto:nkanh@dut.udn.vn}{nkanh@dut.udn.vn}%
}

\date{\today}
\maketitle

\begin{abstract}
Inverter-based resources weaken fault currents in distribution networks, and snapshot-based diagnosis, which classifies a fault from voltage phasors at a few buses at one instant, offers a cheap first-stage screen. Its published accuracies, however, rest on different noise assumptions and do not show whether the error comes from the classifier, the features, the measurement error or the fault resistance. This paper separates these sources on three IEEE test feeders and a public data set with distributed generation, using four feature sets, thirteen classifiers and measurement error defined by total vector error classes. The classifier is not the limiting factor, because the full phasor snapshot classifies almost every fault correctly without noise, and the remaining error comes from measurement error on high-impedance faults. Three voltage sensors classify 90--96\% of faults at the 1\% accuracy class and 96--99\% with distribution micro phasor measurement units (micro-PMUs). A discriminability-margin placement selects them from simulations alone in under one second and matches the best of twelve placement methods. The results show when snapshot-based diagnosis can be trusted.
\end{abstract}


\section{Introduction}\label{sec1}

\begin{nw}
Most interruptions experienced by electricity customers originate in distribution networks, so the speed at which a fault is found and isolated largely determines how long customers remain without supply \cite{stefanidou2022,ngo2025deep}. Utilities report this through the System Average Interruption Duration Index (SAIDI), and faster fault diagnosis is one of the most direct ways to reduce it and to restore supply sooner \cite{stefanidou2022}. The task is becoming harder as distribution networks absorb large shares of inverter-based resources such as photovoltaic systems and battery storage \cite{abedi2026protection}. Unlike synchronous generators, these resources supply a fault current that is limited by their controllers, and their sequence impedances become much larger than those of the lines and the upstream grid \cite{yang2025protection}. Their fault response is governed by the chosen control and current-limiting strategies rather than by physical properties \cite{chakraborty2025influence}, and modes such as momentary cessation can remove a large part of the generation during the disturbance itself \cite{omran2025adaptive}. Fault detection and location are therefore regarded as key open problems in the protection of active distribution networks \cite{abedi2026protection}.

Fault diagnosis covers three related tasks, namely detecting that a fault has occurred, identifying its type, and locating it \cite{stefanidou2022}. Classical approaches solve these tasks from circuit theory \cite{kahnamouei2025mixed,rasoulnia2026comprehensive}. Impedance-based methods estimate the fault distance from the fundamental-frequency voltages and currents recorded at the substation before and during the fault \cite{moraflorez2008}. Model-based methods compare a small number of measurements with the response of a network model to candidate faults, which allows them to work with sparse measurements \cite{jamali2016,lv2023}. Dedicated schemes have also been developed for difficult cases, such as the detection of high-impedance ground faults with transition resistances of several thousand ohms \cite{liang2026high} and the tracking of random arc faults whose behaviour changes during the event \cite{liu2026real}. These methods are interpretable, but they depend on accurate network models and parameters, and their accuracy falls when the fault current is small, either because the fault resistance is high or because inverters limit their contribution \cite{yang2025protection}.

The growing number of synchronised measurement devices in distribution networks, including phasor measurement units (PMUs), smart meters and waveform recorders, has extended the information available for diagnosis \cite{rasoulnia2026comprehensive}. These measurements are not free of error, because distribution waveforms contain more noise, harmonics and phase unbalance than transmission waveforms, which makes phasor estimation more difficult \cite{pandey2025novel}. Because devices cannot be installed at every bus, the choice of their locations has become part of the diagnosis problem \cite{anguswamy2025holistic}. Classical placement methods aim at network observability, for example through integer linear programming (ILP) models that exploit zero-injection buses \cite{chao2026phased}. More recent methods choose locations for diagnosis directly \cite{kaliuzhnyi2027statistical,khattak2025}. One study places PMUs by maximising the smallest Mahalanobis distance between pairs of fault scenarios and links this distance to the Bayes error, so that the feasibility of classification can be assessed before any model is trained \cite{kaliuzhnyi2027statistical}.

Machine learning and artificial intelligence form the most recent step of this development \cite{oelhaf2025}. Instead of relying on a network model, they learn the relation between measurements and fault conditions from data, and they have reported high accuracy with increasingly expressive models \cite{ngo2025deep,oelhaf2025}. Graph attention networks applied to micro-PMU measurements detect, classify and locate faults jointly \cite{ngo2025deep}. A heterogeneous multi-task graph network on the IEEE 123-node feeder estimates the fault type, location, resistance and current from voltage phasors \cite{chanda2025heterogeneous}, and residual graph networks reconstruct the features of unmeasured buses when measurements are sparse \cite{sun2026fault}. One-dimensional convolutional networks classify eleven fault types from short sequences of PMU measurements at selected buses \cite{khattak2025}, wavelet features of feeder-head currents give near-perfect classification on a radial feeder \cite{mamuya2020}, and optimal placement of distribution PMUs has been combined with learned classifiers and locators on the IEEE 13-, 34- and 123-node feeders \cite{anguswamy2025holistic}. Large synthetic data sets have also been released to benchmark such models across many feeder topologies \cite{chen2026large}.

Three limitations remain in this line of work. First, the high accuracies are usually obtained with demanding inputs, either a time window of samples or a measurement at every node; the graph model in \cite{chanda2025heterogeneous} builds its features from all 128 nodes, and the benchmark in \cite{chen2026large} assumes that every node is observable. Second, measurement error is treated inconsistently. A review of fault classification methods notes that measurement noise is frequently ignored in their evaluation \cite{stefanidou2022}, and studies that include it use measures that cannot be compared with one another or with the accuracy classes of PMU standards \cite{ieeec37118}, such as a signal-to-noise ratio \cite{zhang2025data}, a percentage level \cite{ngo2025deep} or a total vector error (TVE) \cite{sun2026fault}. A recent standardised evaluation framework for machine learning in protection makes the reporting of measurements and their degradation mandatory, but it represents noise only as a signal-to-noise ratio on waveforms \cite{oelhaf2026standardized}. Third, the reported accuracy is rarely decomposed into the parts due to the model, the features, the measurement error and the fault resistance \cite{oelhaf2025,sun2026fault}. It is therefore unclear which of these limits the diagnosis, which investment would remove the limit, and in which conditions the method can be trusted \cite{oelhaf2026standardized,oelhaf2027architectures}. From a reliability-engineering perspective, a diagnostic tool can only be used in service when its operating envelope is known and when it can recognise the cases in which it is likely to fail \cite{oelhaf2026standardized,oelhaf2025}.

Table~\ref{tab_lit} compares the closest recent studies with the present work. Its columns state whether the model uses a single-instant (snapshot) input, how many buses are measured, whether the noise model follows the accuracy classes of measurement standards, the largest fault resistance used in training, whether sensors are placed for diagnosis rather than for observability, whether the error is decomposed into its sources, and whether the model can abstain. The graph model in \cite{chanda2025heterogeneous} is trained with all 128 nodes measured, and sparse sets of 10--75\% of the nodes are examined only after training. In \cite{ngo2025deep}, micro-PMUs are installed at all 13 buses of the first test system and at 51 primary buses of the IEEE 123-node feeder. In \cite{sun2026fault}, the model is trained with fault resistances up to 100~$\Omega$ and its generalisation is evaluated up to 2000~$\Omega$ on test data only. In \cite{anguswamy2025holistic}, 37 distribution PMUs cover 26\% of the nodes of the IEEE 123-node feeder. Among the compared studies, only \cite{chanda2025heterogeneous} trains with fault resistances above 100~$\Omega$, and \cite{kaliuzhnyi2027statistical} and \cite{anguswamy2025holistic} also estimate the fault distance, which the present work does not attempt.
\end{nw}

\begin{table}[!htbp]
\centering
\caption{\hlnew{Comparison with recent data-driven fault diagnosis studies in distribution networks.}}\label{tab_lit}
\footnotesize
\renewcommand{\arraystretch}{1.2}
\setlength{\tabcolsep}{4pt}
\begin{tabular*}{\textwidth}{@{\extracolsep{\fill}}lccccccc@{}}
\toprule
 & \multicolumn{2}{c}{Measurement input} & \multicolumn{2}{c}{Test conditions} & \multicolumn{3}{c}{Method}\\
\cmidrule(lr){2-3}\cmidrule(lr){4-5}\cmidrule(l){6-8}
Study & Snapshot & Measured & Standard & $R_f$ max & Diagnostic & Error & Reject\\
      & input    & buses    & noise    & ($\Omega$) & placement & decomp. & option\\
\midrule
Kaliuzhnyi \cite{kaliuzhnyi2027statistical} & -- & 1--3 & \checkmark & 10 & \checkmark & partly & --\\
Ngo \cite{ngo2025deep} & -- & 13--51 & -- & 10 & -- & -- & --\\
Chanda \cite{chanda2025heterogeneous} & \checkmark & all (128) & -- & 500 & partly & -- & --\\
Sun \cite{sun2026fault} & \checkmark & 10--23 & \checkmark & 100 & -- & -- & --\\
Khattak \cite{khattak2025} & -- & 5--9 & -- & 10 & \checkmark & -- & --\\
Anguswamy \cite{anguswamy2025holistic} & -- & 37 & -- & 50 & -- & -- & --\\
Chen \cite{chen2026large} & \checkmark & all & -- & 10 & -- & -- & --\\
\midrule
\textbf{Proposed} & \checkmark & \textbf{2--9} & \checkmark & 100 & \checkmark & \checkmark & \checkmark\\
\bottomrule
\end{tabular*}
\end{table}

\begin{nw}
This paper addresses these limitations for snapshot-based fault diagnosis, which classifies a fault from the fundamental-frequency voltage phasors at a single instant, measured at a few buses. High reporting rates place non-negligible communication, storage and computation requirements on the monitoring system \cite{shen2026dual}, and adjustable reporting rates have been proposed to reduce that burden \cite{pandey2025novel}. A single-instant input reduces the volume of data that must be reported for each event, which makes this setting attractive as a first-stage screening step before slower and more detailed methods are applied. Whether the reporting rate can be lowered in service depends on the application, because fault identification is among the applications that call for faster reporting rates \cite{pandey2025novel}. The study asks where the error of snapshot-based diagnosis comes from and what removes it. Only voltage phasors are used, because voltage sensors are the cheapest synchronised measurement to add at distribution buses; the use of current measurements is left to future work. Fault type and faulted zone are classified on the IEEE 13-, 34- and 123-node feeders and on the public DER-Fault data set \cite{derfault} of a network with distributed energy resources (DERs), with four feature sets of increasing information content, thirteen classifiers, measurement error from the limit of IEEE C37.118.1 down to the level of distribution micro-PMUs \cite{upmu}, and fault resistances up to 100~$\Omega$. The contributions are as follows.
\begin{enumerate}
\item A framework for modelling and reporting phasor measurement error in data-driven fault diagnosis, applicable to any class of PMU. It defines accuracy classes by a percentile of the TVE, converts the measures used in earlier studies to a common TVE scale where this is possible, and shows that the common reading of 1\% TVE as a standard deviation simulates a device that fails the standard and changes the conclusions. 
\item An error decomposition showing that the classifier is not the limiting factor, together with a detectability limit on fault resistance that depends on the measurement error and on the transfer impedance between the fault and the sensors. The limit is turned into the measurement accuracy required for a target diagnostic accuracy and fault-resistance range. 
\item A discriminability-margin placement (DMP), solved exactly as an ILP from simulations alone, with an optional constraint that keeps fault classes separable after the loss of any single sensor. DMP is compared with twelve placement methods, including the distinguishability-based method of \cite{kaliuzhnyi2027statistical}, for both fault type and faulted zone. 
\item An abstention mechanism (reject option) that lets the diagnosis abstain on the snapshots it is likely to misclassify, based either on the classifier probability or on a classifier-free margin in noise units. Deferring 4--16\% of the snapshots raises the accuracy on the rest to 99\%, and the deferred snapshots are mainly high-impedance faults, which turns the operating envelope into a quantity that can be used when deciding whether the method can be trusted as a first-stage screening tool.
\end{enumerate}

The remainder of this paper is organised as follows. Section~\ref{sec2} describes the fault simulation and snapshot extraction, section~\ref{sec3} presents the feature sets, the classifiers and the sensor placement, section~\ref{sec4} defines the experiments, section~\ref{sec5} reports the results, and section~\ref{sec6} concludes the paper.
\end{nw}

\section{Fault data generation and snapshot extraction}\label{sec2}

\subsection{Distribution network simulation environment}\label{subsec2_1}

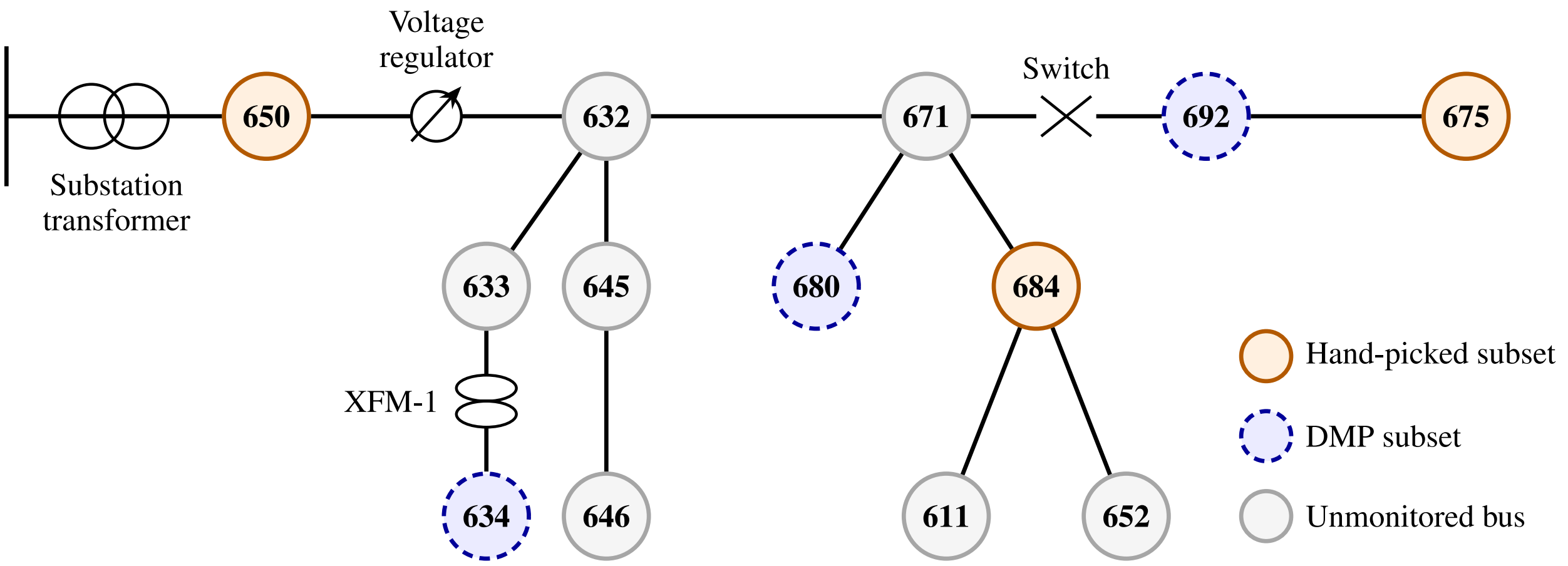
\begin{figure}[t]
\centering
\resizebox{\textwidth}{!}{%
\begin{tikzpicture}[
  font=\small,
  bus/.style={circle, draw=gray!70, very thick, fill=gray!8, minimum size=8.5mm, inner sep=0pt, text=black, font=\small\bfseries},
  obs/.style={bus, draw=orange!70!black, fill=orange!12},
  dmp/.style={bus, draw=blue!60!black, fill=blue!8, densely dashed},
  ln/.style={black, very thick},
  note/.style={font=\footnotesize, text=black, align=center}
]
\node[obs] (650) at (0,0) {650};
\node[bus] (632) at (3.4,0) {632};
\node[bus] (671) at (6.6,0) {671};
\node[dmp] (692) at (9.4,0) {692};
\node[obs] (675) at (12.0,0) {675};
\node[bus] (633) at (2.2,-1.7) {633};
\node[dmp] (634) at (2.2,-4.0) {634};
\node[bus] (645) at (3.4,-1.7) {645};
\node[bus] (646) at (3.4,-4.0) {646};
\node[dmp] (680) at (5.5,-1.7) {680};
\node[obs] (684) at (7.7,-1.7) {684};
\node[bus] (611) at (6.8,-4.0) {611};
\node[bus] (652) at (8.6,-4.0) {652};
\draw[ln] (-2.6,0.7) -- (-2.6,-0.7);
\draw[ln] (-2.6,0) -- (650);
\draw[black, thick] (-1.75,0) circle (0.32) (-1.3,0) circle (0.32);
\node[note, below=0.45cm] at (-1.5,0) {Substation\\transformer};
\draw[ln] (650) -- (632);
\draw[black, thick, fill=white] (1.7,0) circle (0.25);
\draw[-{Stealth}, black, thick] (1.45,-0.25) -- (1.95,0.3);
\node[note, above=0.3cm] at (1.7,0) {Voltage\\regulator};
\draw[ln] (632) -- (671);
\draw[ln] (671) -- (7.7,0);
\draw[ln] (8.3,0) -- (692);
\draw[black, thick] (7.75,-0.2) -- (8.25,0.2) (7.75,0.2) -- (8.25,-0.2);
\node[note, above=0.25cm] at (8.0,0) {Switch};
\draw[ln] (692) -- (675);
\draw[ln] (632) -- (633);
\draw[ln] (632) -- (645) -- (646);
\draw[ln] (671) -- (680);
\draw[ln] (671) -- (684);
\draw[ln] (684) -- (611);
\draw[ln] (684) -- (652);
\draw[ln] (633) -- (2.2,-2.6);
\draw[ln] (2.2,-3.1) -- (634);
\draw[black, thick] (2.2,-2.72) ellipse (0.3 and 0.13) (2.2,-2.98) ellipse (0.3 and 0.13);
\node[note, left=0.35cm] at (2.2,-2.85) {XFM-1};
\node[obs, minimum size=5mm, label={[note]right:Hand-picked subset}] at (10.0,-2.4) {};
\node[dmp, minimum size=5mm, label={[note]right:DMP subset}] at (10.0,-3.2) {};
\node[bus, minimum size=5mm, label={[note]right:Unmonitored bus}] at (10.0,-4.0) {};
\end{tikzpicture}}
\caption{IEEE 13-node test feeder with the hand-picked (shaded) and DMP (dashed) three-bus subsets.}\label{fig1}
\end{figure}

\begin{nw}
Fault scenarios are simulated using the Open Distribution System Simulator (OpenDSS) \cite{dugan2011}, a widely adopted distribution system simulation tool for unbalanced networks. OpenDSS is selected due to its ability to model three-phase voltage phasors, single- and two-phase laterals, and network asymmetries with engineering-level fidelity.

Three IEEE radial test feeders are used: the 13-node feeder (14 buses, figure~\ref{fig1}), the 34-node feeder (34 buses) and the 123-node feeder (127 buses including regulator and switch nodes). The previous IEEE 118-bus transmission case, built from a bus impedance matrix, is replaced by these feeders because a balanced $Z_{\mathrm{bus}}$ model cannot represent the unbalanced, single-phase laterals that dominate distribution faults. As an external check, the public DER-Fault data set \cite{derfault} is also used: 12\,600 fault events on the Iowa 240-bus system under six DER scenarios (S1 without DER to S6 with the highest penetration), with 93 PMU-capable buses.
\end{nw}

\subsection{Fault data generation and snapshot construction}\label{subsec2_2}

\begin{figure}[t]
\centering
\begin{nwf}
\centering
\begin{tikzpicture}[
  font=\footnotesize,
  node distance=5mm,
  main/.style={draw, line width=0.6pt, fill=white, align=left, text width=6.1cm, inner sep=4pt},
  side/.style={draw, line width=0.6pt, fill=gray!8, align=left, text width=4.4cm, inner sep=4pt},
  feat/.style={draw, line width=0.5pt, fill=white, align=center, text width=1.27cm, minimum height=1.0cm, inner sep=2pt},
  arr/.style={-{Stealth[length=2mm]}, line width=0.6pt},
  hd/.style={font=\footnotesize\bfseries},
  stage/.style={font=\footnotesize\bfseries, rotate=90, anchor=south}
]
\node[main] (net) {\textbf{Test networks}\\
  IEEE 13-, 34- and 123-node feeders (OpenDSS)\\
  DER-Fault: Iowa 240-bus system, DER scenarios S1--S6};
\node[main, below=of net] (scen) {\textbf{Random fault scenario}\\
  Type: SLG, LL, 3PH (LLG for the four-class task)\\
  Faulted bus and phases drawn uniformly\\
  $R_f$ log-uniform in $[0.001,\,100]~\Omega$; random load scaling};
\node[main, below=of scen] (sim) {\textbf{Phasor solution}\\
  Pre-fault power flow $\rightarrow$ $V^{\mathrm{pre}}_{\phi,b}$\\
  Fault study $\rightarrow$ $V^{\mathrm{fault}}_{\phi,b}$ at every node};
\node[main, below=of sim] (pmu) {\textbf{PMU measurement at $k$ monitored buses}\\
  Independent phasor noise on $V^{\mathrm{pre}}$ and $V^{\mathrm{fault}}$:\\
  1\% TVE ($\sigma_m=1\%$, $\sigma_\theta=0.5^\circ$) or\\
  micro-PMU ($\sigma_m=0.02\%$, $\sigma_\theta=0.002^\circ$)};
\node[feat, below=7mm of pmu.south west, anchor=north west] (f1) {\textbf{AGG}\\$f_1$--$f_4$\\(4)};
\node[feat, right=1.7mm of f1] (f2) {\textbf{PPA}\\per phase\\(12)};
\node[feat, right=1.7mm of f2] (f3) {\textbf{SEQ}\\dips, $z_0$, $z_2$\\(13)};
\node[feat, right=1.7mm of f3] (f4) {\textbf{FPS}\\ratio, angle\\($2n+2n_b$)};
\node[draw, densely dashed, line width=0.5pt, fit=(f1)(f4), inner sep=3pt,
      label={[hd, anchor=south west, inner sep=1pt]north west:Feature sets}] (fs) {};
\node[main, below=7mm of fs.south, anchor=north] (clf) {\textbf{Classification}\\
  13 classifiers and a label-free nearest-template classifier\\
  Stratified five-fold cross-validation (two seeds)};
\node[main, below=of clf] (ev) {\textbf{Evaluation}\\
  Accuracy, macro-F1, HIF ($R_f>10~\Omega$) and N-1 accuracy, ECE\\
  Friedman and Wilcoxon tests; Cover--Hart bound\\
  Load shift, noise mismatch, PMU loss, transfer};
\draw[arr] (net) -- (scen);
\draw[arr] (scen) -- (sim);
\draw[arr] (sim) -- (pmu);
\draw[arr] (pmu.south) -- (pmu.south |- fs.north);
\draw[arr] (fs.south) -- (clf.north);
\draw[arr] (clf) -- (ev);
\coordinate (bx) at ($(net.east)+(11mm,0)$);
\node[side, anchor=west] (grid) at (bx |- net.east) {\textbf{Design grid}\\
  Every bus $\times$ fault type $\times$ phases $\times$ eight $R_f$ values (0.01--100~$\Omega$)};
\node[side, anchor=west] (mc) at (bx |- scen.east) {\textbf{Noise statistics}\\
  40 Monte-Carlo draws of 1\% TVE: signature $\mathbf{s}_b$ and spread $\boldsymbol{\sigma}_b$ at every bus};
\node[side, anchor=west] (mg) at (bx |- sim.east) {\textbf{Margin} $m_{eb}$, eq.~(\ref{eq8})\\
  Bus $b$ covers event $e$ if $m_{eb}\ge3$};
\node[side, anchor=west] (ilp) at (bx |- pmu.east) {\textbf{Max-coverage ILP}, eq.~(\ref{eq9})\\
  Choose $k$ buses; N-1 variant requires two covering buses};
\node[draw, densely dashed, line width=0.5pt, fit=(grid)(ilp), inner sep=3pt,
      label={[hd, anchor=south west, inner sep=1pt]north west:DMP placement (offline, no labels)}] (dmp) {};
\draw[arr] (net.east) -- (grid.west);
\draw[arr] (grid) -- (mc);
\draw[arr] (mc) -- (mg);
\draw[arr] (mg) -- (ilp);
\draw[arr] (ilp.west) -- node[above, font=\scriptsize] {buses} (pmu.east);
\draw[line width=0.5pt] ([xshift=-3mm]net.north west) -- ([xshift=-3mm]sim.south west)
  node[midway, stage, xshift=0mm, yshift=1mm] {Simulation};
\draw[line width=0.5pt] ([xshift=-3mm]pmu.north west) -- ([xshift=-3mm]fs.south west)
  node[midway, stage, yshift=1mm] {Measurement};
\draw[line width=0.5pt] ([xshift=-3mm]clf.north west) -- ([xshift=-3mm]ev.south west)
  node[midway, stage, yshift=1mm] {Learning};
\end{tikzpicture}
\caption{Workflow of data generation, sensor placement, measurement and evaluation.}\label{fig2}
\end{nwf}
\end{figure}
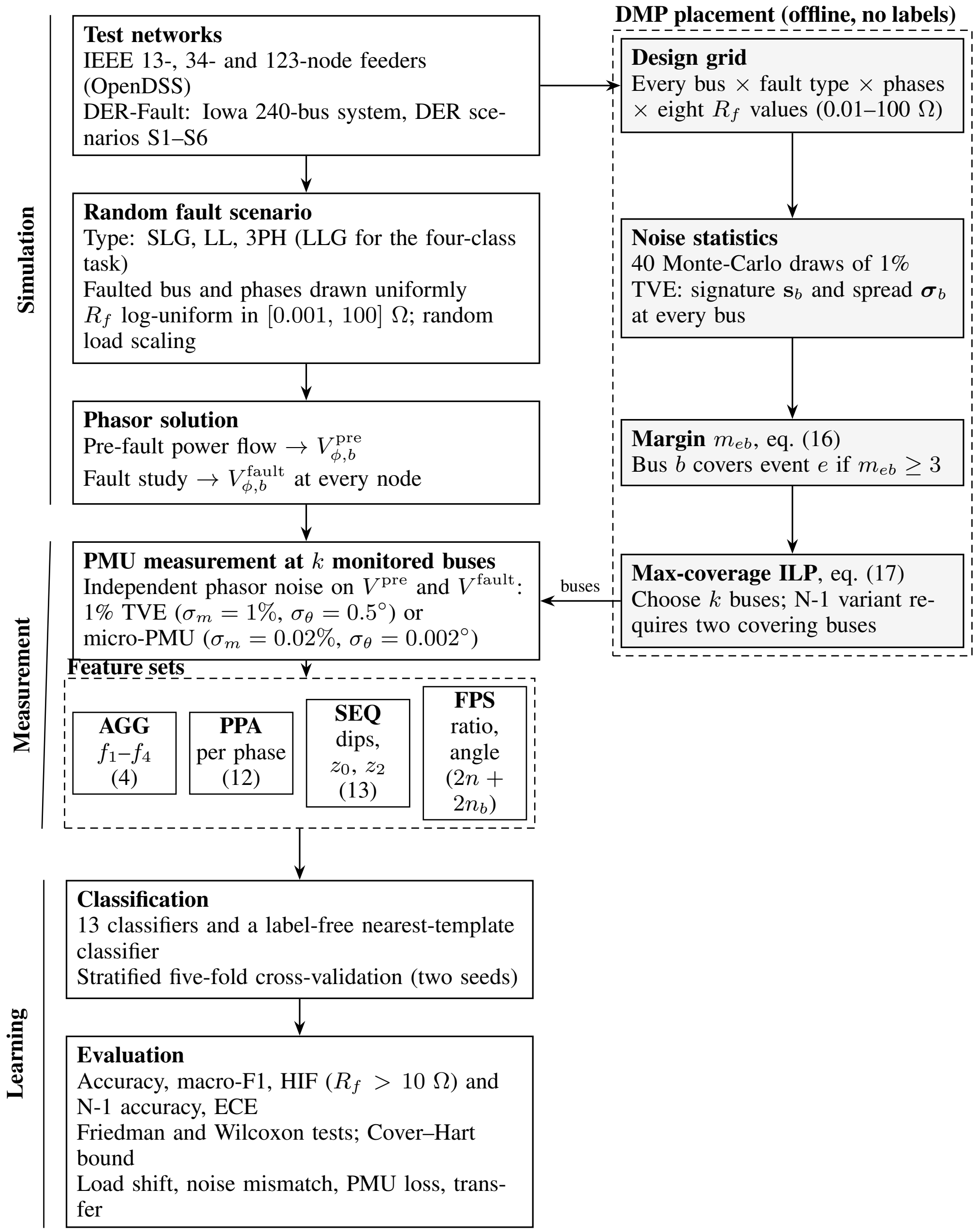

\begin{nw}
Fault data are generated following the workflow in figure~\ref{fig2}. Each simulation solves the pre-fault power flow, applies a fault with randomly selected type, location, faulted phases and resistance, and solves the faulted network in the OpenDSS fault-study mode. Loads are scaled randomly in every simulation to vary the operating point.

Fault resistance $R_f$ is sampled log-uniformly from 0.001~$\Omega$ (bolted faults) to 100~$\Omega$ (high-impedance faults). Each data set contains about 9\,000 balanced samples of single-line-to-ground (SLG), line-to-line (LL) and three-phase (3PH) faults (8\,944, 9\,060 and 8\,961 samples for the 13-, 34- and 123-node feeders). Line-to-line-to-ground faults are simulated for the four-class task of section~\ref{subsec5_8}.
\end{nw}

\begin{nw}
\textit{Measurement noise.} The previous $\pm$2\% perturbation is replaced by phasor measurement error applied independently to the pre-fault and fault phasors of every monitored node, following the framework of section~\ref{subsec3_err}. Two reference classes are used. The first meets the 1\% TVE limit of IEEE C37.118.1 \cite{ieeec37118} at the 99th percentile, with $\sigma_m=0.33\%$ and $\sigma_\theta=0.19^\circ$. The second uses the noise measured on distribution-level micro-PMUs \cite{upmu}, $\sigma_m=0.02\%$ and $\sigma_\theta=0.002^\circ$. Results are reported for the 1\% TVE accuracy class unless stated otherwise.
\end{nw}

The system response is simulated over a short interval to capture the immediate electrical impact. Three-phase bus voltages are sampled at a fixed time instant shortly after fault inception, representing the faulted state without time-series information. Voltage snapshots are then subsampled for sparsity analysis, normalised to pre-fault values, and converted into feature vectors. Repeating across fault scenarios yields a labelled dataset for controlled evaluation.

\section{Feature design and classifiers}\label{sec3}

\begin{nw}
\subsection{Snapshot measurement model}\label{subsec3_0}

Let $\mathcal{O}$ be the set of monitored buses and $\mathcal{I}$ the set of $n$ monitored nodes (bus--phase pairs), and let $V^{\mathrm{pre}}_i$ and $V^{\mathrm{fault}}_i$ be the true complex voltage phasors of node $i\in\mathcal{I}$ before and during the fault. A PMU reports
\begin{equation}\label{eq_meas}
\hat V^{\mathrm{t}}_i = V^{\mathrm{t}}_i\,(1+\varepsilon^{\mathrm{t}}_{m,i})\,e^{\,j\varepsilon^{\mathrm{t}}_{\theta,i}},\qquad
\varepsilon^{\mathrm{t}}_{m,i}\sim\mathcal{N}(0,\sigma_m^2),\quad \varepsilon^{\mathrm{t}}_{\theta,i}\sim\mathcal{N}(0,\sigma_\theta^2),
\end{equation}
for $\mathrm{t}\in\{\mathrm{pre},\mathrm{fault}\}$, where $j$ is the imaginary unit, $\varepsilon_m$ and $\varepsilon_\theta$ are the magnitude and angle errors, and $\sigma_m$ and $\sigma_\theta$ are their standard deviations (section~\ref{subsec2_2}); all errors are mutually independent. The snapshot available to the classifier is $\mathbf{X}=\{(\hat V^{\mathrm{pre}}_i,\hat V^{\mathrm{fault}}_i)\}_{i\in\mathcal{I}}$, and the task is to estimate the fault class $c\in\mathcal{C}=\{\mathrm{SLG},\mathrm{LL},\mathrm{3PH}\}$ from $\mathbf{X}$ or from a feature map $\mathbf{x}=g(\mathbf{X})$.

The fault leaves its signature in the complex ratio $\rho_i=V^{\mathrm{fault}}_i/V^{\mathrm{pre}}_i$, whose measured value $\hat\rho_i$ is formed from the reported phasors. To first order in the noise, its measured magnitude and angle are
\begin{equation}\label{eq_ratio}
|\hat\rho_i|\approx|\rho_i|\big(1+\varepsilon^{\mathrm{fault}}_{m,i}-\varepsilon^{\mathrm{pre}}_{m,i}\big),\qquad
\angle\hat\rho_i=\angle\rho_i+\varepsilon^{\mathrm{fault}}_{\theta,i}-\varepsilon^{\mathrm{pre}}_{\theta,i},
\end{equation}
so the voltage sag $d_i=1-|\rho_i|$ of node $i$ is observed with standard deviation $\sqrt2\,\sigma_m|\rho_i|$ and the angle shift with standard deviation $\sqrt2\,\sigma_\theta$. Normalising by the pre-fault phasor removes the load-dependent operating point but doubles the noise variance.
\end{nw}

\subsection{A reporting framework for phasor measurement error}\label{subsec3_err}

\begin{nw}
The studies in table~\ref{tab_lit} express measurement error in different and mostly incompatible ways, so their accuracies cannot be compared or related to a real device. This paper therefore uses one framework for all experiments, built on the total vector error (TVE) of IEEE C37.118.1 \cite{ieeec37118}. For a reported phasor $\hat V=V(1+\varepsilon_m)e^{j\varepsilon_\theta}$,
\begin{equation}\label{eq_tve}
\mathrm{TVE}=\frac{|\hat V-V|}{|V|}\approx\sqrt{\varepsilon_m^2+\varepsilon_\theta^2}
\end{equation}
for small errors, with $\varepsilon_\theta$ in radians. The standard gives 1\% TVE as a limit that a compliant device must not exceed, not as a standard deviation. A device is therefore said to belong to the $x$\% TVE accuracy class when the 99th percentile of its TVE equals $x$\%. When the error is shared equally between magnitude and angle ($\sigma_m=\sigma_\theta=\sigma$, $\sigma_\theta$ in radians), $\mathrm{TVE}/\sigma$ follows a Rayleigh distribution and
\begin{equation}\label{eq_tveclass}
\sigma=\frac{x}{\sqrt{-2\ln(1-p)}}=\frac{x}{3.035}\quad\text{for } p=0.99 .
\end{equation}
A synchronisation error with standard deviation $\Delta t$ adds an angle error $2\pi f\Delta t$ in quadrature, i.e.\ $0.0216^\circ$ per microsecond at 60~Hz. Table~\ref{tab_noise} lists the reference levels used in this paper. Reading the 1\% limit as $\sigma_m=1\%$ with $\sigma_\theta=0.5^\circ$, as is common, gives a device whose TVE exceeds 1\% in 56\% of the measurements and whose 99th percentile is 2.9\%; it is practically a 3\% TVE device.

Three rules follow for reporting. First, the error is applied to the raw phasors before any feature is computed, because an error added to standardised features depends on the data set and cannot be related to a device. Second, the class is stated as a percentile of the TVE together with $\sigma_m$ and $\sigma_\theta$. Third, a signal-to-noise ratio on waveforms is reported together with the length of the phasor estimation window, because the same ratio gives very different phasor errors for different windows; percentages without a stated basis cannot be converted to TVE.
\end{nw}

\begin{table}[!htbp]
\centering
\caption{\hlnew{Reference levels of phasor measurement error. TVE statistics (\%) and share of measurements above 1\% TVE (\%), from $2\times10^5$ draws.}}\label{tab_noise}
\small
\begin{tabular}{lccccc}
\toprule
Level & $\sigma_m$ (\%) & $\sigma_\theta$ ($^\circ$) & Median TVE & 99th pct.\ TVE & TVE $>1\%$ \\
\midrule
Micro-PMU (measured \cite{upmu}) & 0.02 & 0.002 & 0.01 & 0.05 & 0\\
1\% TVE accuracy class & 0.33 & 0.19 & 0.39 & 1.00 & 1\\
3\% TVE accuracy class & 0.99 & 0.57 & 1.16 & 3.00 & 60\\
Common reading of 1\% TVE & 1.00 & 0.50 & 1.10 & 2.86 & 56\\
\bottomrule
\end{tabular}
\end{table}

\begin{nw}
\subsection{Fault signature and fault resistance}\label{subsec3_0b}

The size of the signature follows from the sequence networks seen from the faulted bus $q$. With $Z^{(0)}_{qq}$, $Z^{(1)}_{qq}$ and $Z^{(2)}_{qq}$ the zero-, positive- and negative-sequence driving-point impedances, $V_q$ the pre-fault voltage at bus $q$ and $R_f$ the fault resistance, the fault currents $I_f$ of the three classes are
\begin{equation}\label{eq_if}
I_f^{\mathrm{SLG}}=\frac{3V_q}{Z^{(0)}_{qq}+Z^{(1)}_{qq}+Z^{(2)}_{qq}+3R_f},\qquad
|I_f^{\mathrm{LL}}|=\frac{\sqrt3\,|V_q|}{|Z^{(1)}_{qq}+Z^{(2)}_{qq}+R_f|},\qquad
I_f^{\mathrm{3PH}}=\frac{V_q}{Z^{(1)}_{qq}+R_f}.
\end{equation}
The voltage change at a monitored bus $b$ is the transfer impedance $Z_{bq}$ times this current, so the sag seen at $b$, $d_b=\max_{i\in b}d_i$ over the phases of bus $b$, satisfies $d_b\le\kappa_{bq}\,|V_q|/|Z_{qq}+\alpha R_f|$, with $\kappa_{bq}$ the largest sequence transfer impedance between $b$ and $q$, $Z_{qq}$ the corresponding sum of driving-point impedances, and $\alpha=3,1,1$ for the three classes. For $R_f\gg|Z_{qq}|$ the sag therefore decays as
\begin{equation}\label{eq_dip}
d_b(R_f)\approx\frac{\kappa_{bq}}{R_f}\qquad(\text{per unit, }|V_q|\approx1),
\end{equation}
and the signatures of all three classes converge to the pre-fault steady state, $\rho_i\to1$. By equation~(\ref{eq_ratio}), two classes can be separated at bus $b$ only while their sag difference exceeds a few noise standard deviations, $|d_b^{c}-d_b^{c'}|\ge\tau\sqrt2\,\sigma_m$, where $d_b^{c}$ is the sag produced by a fault of class $c$ and $\tau$ is a separation factor ($\tau=3$ in section~\ref{subsec3_4}). Combining this with equation~(\ref{eq_dip}) gives a detectability limit of the form
\begin{equation}\label{eq_rfstar}
R_f^{\star}\;\propto\;\frac{\kappa_{bq}}{\tau\sqrt2\,\sigma_m},
\end{equation}
above which snapshot classification at bus $b$ degrades towards chance. Two consequences are tested in section~\ref{sec5}: the accuracy loss should be concentrated in the highest $R_f$ bins, and $R_f^{\star}$ should grow in inverse proportion to the noise level, so that reducing $\sigma_m$ from 1\% to the micro-PMU level should recover most high-impedance faults. Equation~(\ref{eq_rfstar}) also shows why sensor placement matters: buses with a large transfer impedance $\kappa_{bq}$ to many fault locations extend the detectable range of $R_f$.
\end{nw}

\subsection{Snapshot-based feature construction}\label{subsec3_1}

Each data sample corresponds to a single time instant captured during a fault event. \hlnew{Instead of using full transient waveforms, only the three-phase voltage phasors at the monitored buses are used,} reflecting realistic measurement constraints in distribution networks. A snapshot is extracted shortly after fault inception, when voltage deviations have fully developed but before protection actions take place.

Let $V_{\phi,b}^{\mathrm{fault}}$ denote the per-unit voltage magnitude of phase $\phi \in \{\mathrm{a},\mathrm{b},\mathrm{c}\}$ at bus $b$ during the fault snapshot, and $V_{\phi,b}^{\mathrm{pre}}$ the corresponding pre-fault voltage. The voltage deviation vector is
\begin{equation}\label{eq2}
\Delta V = V^{\mathrm{fault}} - V^{\mathrm{pre}}.
\end{equation}
Four scalar features are computed from the snapshot:
\begin{align}
f_{1} &= \mathrm{mean}\left(V^{\mathrm{fault}}\right), \label{eq3}\\
f_{2} &= \mathrm{std}\left(V^{\mathrm{fault}}\right), \label{eq4}\\
f_{3} &= \max\left(\left|\Delta V\right|\right), \label{eq5}\\
f_{4} &= \left\Vert \Delta V \right\Vert^{2}, \label{eq6}
\end{align}
namely the mean fault voltage, its standard deviation, the maximum absolute voltage deviation, and the voltage deviation energy. These features capture complementary aspects of fault-induced disturbances and form a low-dimensional, network-size-independent representation suitable for controlled benchmarking.

\begin{nw}
The four statistics pool all phases and buses, so they discard which phase is depressed and the phase angle of each voltage. To measure how much accuracy this costs, three further feature sets are defined, each adding one kind of information:
\begin{itemize}
\item \textbf{AGG} (aggregate features, 4 features): $f_1$--$f_4$ of equations~(\ref{eq3})--(\ref{eq6}).
\item \textbf{PPA} (per-phase aggregate features, 12 features): $f_1$--$f_4$ computed separately for phases a, b and c.
\item \textbf{SEQ} (sequence-component features, 13 features): the three per-phase voltage sags $d_\phi=1-|\hat\rho_\phi|$, with $\hat\rho_\phi$ the measured ratio of phase $\phi$, at the most disturbed monitored bus $b^\ast=\arg\max_{b\in\mathcal{O}}\sum_\phi d_{\phi,b}$, sorted in descending order; the maximum and mean over monitored buses of the zero- and negative-sequence ratios
\begin{equation}\label{eq7}
z_0=\frac{|V_0|}{|V_1|},\qquad z_2=\frac{|V_2|}{|V_1|},\qquad
\begin{bmatrix}V_0\\V_1\\V_2\end{bmatrix}=\frac13\begin{bmatrix}1&1&1\\1&\lambda&\lambda^2\\1&\lambda^2&\lambda\end{bmatrix}\begin{bmatrix}\hat V^{\mathrm{fault}}_{\mathrm{a}}\\\hat V^{\mathrm{fault}}_{\mathrm{b}}\\\hat V^{\mathrm{fault}}_{\mathrm{c}}\end{bmatrix},\quad \lambda=e^{j2\pi/3},
\end{equation}
where $V_0$, $V_1$ and $V_2$ are the zero-, positive- and negative-sequence components and $\lambda$ is the $120^\circ$ rotation operator; and of the positive-sequence drop $1-|V_1|/|V_1^{\mathrm{pre}}|$, with $V_1^{\mathrm{pre}}$ computed from the pre-fault phasors; and $f_1$--$f_4$. Sorting makes the sags invariant to which phase is faulted, while $z_0$ and $z_2$ separate the classes by symmetry: in a balanced network $z_0=z_2=0$ for 3PH, $z_0=0<z_2$ for LL, and $z_0>0$ for SLG.
\item \textbf{FPS} (full phasor snapshot, $2n+2n_b$ features for $n$ monitored nodes on $n_b$ three-phase buses): for every monitored node the magnitude ratio $|\hat\rho_i|$ and the angle shift $\angle\hat\rho_i$, plus $z_0$ and $z_2$ of every monitored three-phase bus.
\end{itemize}
The sets are ordered by information content, but they are not nested: SEQ, for example, contains sequence quantities that PPA does not.

\textit{Information content and the Bayes error.} For a feature map $g$, let $\varepsilon^\ast(g)=1-\mathbb{E}\big[\max_{c}P(c\mid g(\mathbf{X}))\big]$ be the Bayes error, the lowest error any classifier can reach with input $g(\mathbf{X})$; here $P(c\mid\cdot)$ is the posterior probability of class $c$ and the expectation is over snapshots. Because conditioning on less information cannot increase the maximum posterior,
\begin{equation}\label{eq_dpi}
\varepsilon^\ast(h\circ g)\ \ge\ \varepsilon^\ast(g)\qquad\text{for every further map } h ,
\end{equation}
AGG, PPA and SEQ are functions of the ratios $\{\hat\rho_i\}$, and FPS contains every $\hat\rho_i$ in polar form, so $\varepsilon^\ast(\mathrm{AGG})$, $\varepsilon^\ast(\mathrm{PPA})$ and $\varepsilon^\ast(\mathrm{SEQ})$ are all bounded below by $\varepsilon^\ast(\mathrm{FPS})$. A gap in accuracy between FPS and a smaller set therefore measures information discarded by the feature map, whereas a gap between classifiers on the same set measures a learning limitation. Without noise, FPS reaches at least 99.8\% accuracy in every test fold (section~\ref{subsec5_0}), which is consistent with $\varepsilon^\ast(\mathrm{FPS})\approx0$; any remaining error under noise is then caused by the noise. With noise, $\varepsilon^\ast$ is estimated from the leave-one-out nearest-neighbour error $\varepsilon_{\mathrm{NN}}$ through the Cover--Hart inequality for $M=|\mathcal{C}|$ classes \cite{coverhart1967,sekeh2020},
\begin{equation}\label{eq_ch}
\varepsilon^\ast\ \ge\ \frac{M-1}{M}\left(1-\sqrt{1-\frac{M}{M-1}\,\varepsilon_{\mathrm{NN}}}\right),
\end{equation}
which gives an upper bound $1-\varepsilon^\ast$ on the accuracy of any classifier.
\end{nw}

\subsection{Fault labelling strategy}\label{subsec3_2}

Each snapshot is labelled according to the simulated fault type: single-line-to-ground (1$\phi$), line-to-line (2$\phi$), or three-phase (3$\phi$). The task is therefore a three-class problem in which the classifier must infer the fault type solely from snapshot voltage features. No information about fault location, impedance, or topology is provided, so performance reflects the intrinsic discriminative power of snapshot voltage measurements. \hlnew{Four-class (with line-to-line-to-ground) and ten-class phase-resolved tasks are evaluated in section~\ref{subsec5_8}.}

\subsection{Baseline classifiers}\label{subsec3_3}

\textit{Support vector machine (SVM).} A multi-class SVM with a radial basis function (RBF) kernel is used. Features are standardised to zero mean and unit variance prior to training.

\textit{Random forest (RF).} An ensemble of decision trees captures nonlinear relationships between features and provides robustness against noise; it operates directly on raw features.

\begin{nw}
To test whether the observed performance ceiling is intrinsic to snapshot features or specific to the classifier, eleven further classifiers are included: logistic regression (LR), $k$-nearest neighbours (kNN) with distance-weighted voting, a multilayer perceptron (MLP) with two hidden layers (64 and 32 neurons), linear discriminant analysis (LDA), quadratic discriminant analysis (QDA), Gaussian naive Bayes (NB), extremely randomised trees (ExtraTrees), two gradient-boosting machines (extreme gradient boosting, XGBoost, and the light gradient-boosting machine, LightGBM \cite{lightgbm,sapountzoglou2020}), a stacking ensemble of RF, LightGBM, SVM and kNN, and a one-dimensional convolutional network (CNN1D). A nearest-template classifier that assigns the class of the nearest noise-free design signature serves as a reference that needs no labelled training data. LightGBM is used as the default classifier in sections~\ref{subsec5_1}--\ref{subsec5_8} because it is among the most accurate and is fast to train.
\end{nw}

\begin{figure}[t]
\begin{nwf}
\centering
\includegraphics[width=\textwidth]{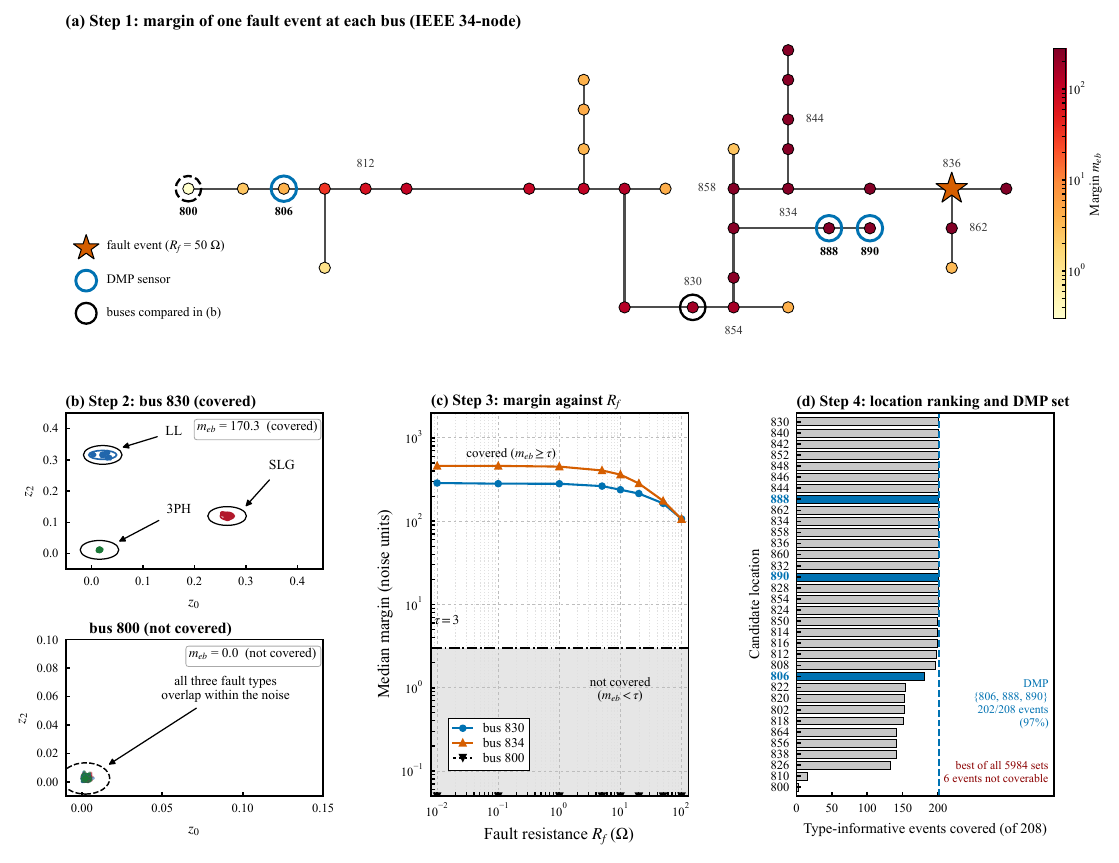}
\caption{\hlnew{Worked example of DMP on the IEEE 34-node feeder for a 50~$\Omega$ fault at bus 836.}}\label{fig_dmp_example}
\end{nwf}
\end{figure}

\begin{nw}
\subsection{Discriminability-margin placement (DMP)}\label{subsec3_4}

The original subset 650--675--684 was chosen by hand. DMP selects sensor sites from simulations only, without labelled data or a trained classifier. A set of design fault scenarios is simulated in which every bus is faulted with every fault type, every feasible set of phases and eight resistances $R_f\in\{0.01,0.1,1,5,10,20,50,100\}~\Omega$. Each pair $e=(\text{fault bus},R_f)$ is called an event; on the 34-node feeder the grid comprises 1\,520 OpenDSS simulations that form 272 events. \hlnew{An event is \emph{type-informative} when at least two fault types are feasible at its bus. Events on single-phase laterals, where only an SLG fault can occur, carry no fault-type information: their margin is set to zero, and they are excluded when coverage is reported.} At every candidate location $b$ a five-element signature $\mathbf{s}_b$ is computed (sorted per-phase voltage sags, $z_0$, $z_2$), and its noise standard deviation $\boldsymbol{\sigma}_b$ is estimated by 40 Monte-Carlo draws of the 1\% TVE noise. Let $\mathbf{s}_b(e,c)$ be the noise-free signature at bus $b$ when event $e$ is a fault of class $c$. The noise-normalised margin of event $e$ at bus $b$ is the smallest distance between signatures of two different fault types,
\begin{equation}\label{eq8}
m_{eb}=\min_{c\neq c'}\left\Vert \frac{\mathbf{s}_b(e,c)-\mathbf{s}_b(e,c')}{\bar{\boldsymbol{\sigma}}_b}\right\Vert_2 ,
\end{equation}
where $\bar{\boldsymbol{\sigma}}_b$ is the pooled standard deviation of the two signatures and the division is element-wise. Bus $b$ covers event $e$ when $m_{eb}\ge\tau$, with $\tau=3$. With binary variables $x_b$ (sensor at $b$) and $y_e$ (event covered), $k$ sensors (PMUs) are placed by the maximum-coverage ILP
\begin{equation}\label{eq9}
\max_{x,y}\ \sum_e y_e \quad\text{s.t.}\quad y_e\le\sum_{b:\,m_{eb}\ge\tau}x_b,\qquad \sum_b x_b=k,\qquad x_b,y_e\in\{0,1\},
\end{equation}
solved exactly with HiGHS. In the N-1 variant, which must tolerate the loss of any single sensor, an event counts only when at least two selected sensors cover it, so that the loss of any one PMU leaves it covered.

The margin is the noise-scaled counterpart of equation~(\ref{eq_rfstar}): $m_{eb}$ is large when the sag differences produced at bus $b$ by event $e$ are large compared with $\sqrt2\,\sigma_m$, so maximising coverage extends the range of $R_f$ over which the selected buses can separate the classes. The objective of equation~(\ref{eq9}) is a monotone submodular function of the sensor set, and the ILP returns its exact maximum.

\hlnew{Figure~\ref{fig_dmp_example} makes the procedure concrete on the 34-node feeder, following a single event (a 50~$\Omega$ fault at bus 836) through the four steps. Panel (a) maps the margin that this event produces at every candidate location. Thirty buses see the event with a margin of 3.3--276 noise standard deviations and therefore cover it; the remaining four see 0.01--2.6 and do not. These are the substation bus 800 and its stub 802, together with the single-phase lateral buses 810 and 864, and they fail for two different reasons: on the laterals the single measured phase carries too little fault-type information, while at the feeder head the substation source holds the voltage close to its pre-fault value, so the residual sag is of the order of the measurement noise. Panel (b) shows the second reason at the level of the signature itself, on axes shared by both panels: at bus 830 the three fault types form well-separated clusters in the $(z_0,z_2)$ plane, whereas at bus 800 all three collapse into a cloud of the size of the noise. Panel (c) shows the $1/R_f$ decay of equation~(\ref{eq_dip}) behind these numbers: the median margin of bus 830 stays above $\tau=3$ at every resistance, whereas that of the feeder head never exceeds 0.03. Panel (d) ranks the candidate locations by the number of type-informative events they cover and marks the set that the ILP selects.}

\end{nw}

\begin{nw}
\hlnew{\textit{What the example exposes.} Two properties of the design fault scenarios bound what any placement method can achieve, and both are visible in figure~\ref{fig_dmp_example}. First, only events at buses with at least two feasible fault classes carry fault-type information. On the 34-node feeder 64 of the 272 design events sit on single-phase laterals, where only an SLG fault can occur and no sensor set can be credited with separating them. Of the remaining 208 events the DMP set covers 202, and an exhaustive search over all 5\,984 three-bus subsets finds no set that covers more. Second, the six events that remain out of reach are all at the substation bus 800, where the three signatures of every candidate location lie inside the noise. These are the locations for which equation~(\ref{eq_rfstar}) predicts failure, and no placement of voltage sensors can recover them; a feeder-head current measurement, which is outside the scope of this paper, would bypass the attenuation that the voltage measurement suffers.}

\textit{Label-free ranking index.} If the signatures of two classes $c,c'$ at bus $b$ were Gaussian with a common diagonal covariance, the pairwise Bayes error of a sensor set $S$ would obey the Bhattacharyya bound $P_{\mathrm{err}}\le\tfrac12\exp(-D^2_{e,S}/8)$, where $D_{e,S}$ is the Mahalanobis distance between the class means of event $e$ over the buses in $S$. With noise independent between PMUs, the covariance is block diagonal over buses and the squared distance adds, $D_{e,S}^2=\sum_{b\in S}m_{eb}^2$. Weighting each event by $w_e$, the share of its $R_f$ level on a logarithmic scale ($\sum_e w_e=1$), gives the index
\begin{equation}\label{eq10}
B(S)=\sum_e w_e\,\tfrac12\exp\!\Big(-\tfrac18\sum_{b\in S}m_{eb}^2\Big).
\end{equation}
Because the weights sum to one, $1-2B(S)=\sum_e w_e\big[1-\exp(-\tfrac18 D_{e,S}^2)\big]$. Each term is a concave, non-decreasing function of the modular quantity $D_{e,S}^2$, so $1-2B(S)$ is monotone non-decreasing and submodular in $S$; minimising $B(S)$ is therefore a monotone submodular maximisation, and greedy selection is within a factor $1-1/e$ of its optimum. Because the signatures are not Gaussian, $B(S)$ is used only to rank candidate sets, not as a bound on the error (section~\ref{subsec5_1}).
\end{nw}

\section{Experimental setup and benchmark protocol}\label{sec4}

\subsection{Benchmark objectives}\label{subsec4_1}

\begin{nw}
Existing works typically evaluate performance under a single configuration. This study defines targeted benchmarks addressing four questions: (i) where the classification error comes from (classifier, features, noise or fault resistance); (ii) how much each kind of snapshot information contributes; (iii) where a small number of sensors should be placed; and (iv) whether the trends hold across classifiers, feeders, operating conditions and DER scenarios.
\end{nw}

\subsection{Benchmark tasks}\label{subsec4_2}

\textit{Measurement sparsity.} Sensing configurations range from all monitored buses to sparse subsets, with feature definitions and hyperparameters kept identical. This assesses whether snapshot-based diagnosis requires dense monitoring, which is rarely available in practice.

\begin{nw}
\textit{Feature sets.} The four feature sets of section~\ref{subsec3_1} are compared with the same classifier and data. The original one-at-a-time removal of $f_1$--$f_4$ is kept as a secondary analysis.

\textit{Sensor placement.} For $k=3$--$9$ sensors, DMP is compared with eleven baselines: random sets, a topology rule (substation and the two mutually most distant leaves), an observability ILP, the monitor-reach-area (MRA) ILP for voltage-sag monitoring \cite{olguin2006,wang2021}, sparse sensor placement by QR factorisation with column pivoting (QR pivot) \cite{manohar2018}, D-optimal greedy selection, mutual-information (MI) greedy selection, information greedy selection (leave-one-out 1-NN error), accuracy greedy selection (wrapper), forward selection with an SVM and neighbour refinement (FSNR-SVM, after \cite{khattak2025}), a genetic algorithm (GA) with a linear discriminant analysis wrapper (GA-LDA), and simulated annealing on the DMP coverage objective. On the 13-node feeder, all 364 three-bus subsets are also evaluated exhaustively. On DER-Fault, the NSGA-II placements published with the data set \cite{derfault,deb2002,andreoni2021} are added.
\end{nw}

\textit{Cross-model and cross-seed robustness.} Classifiers are evaluated under identical data splits and experiments are repeated across multiple random seeds, to determine whether trends reflect intrinsic properties of the features or favourable splits.

\begin{nw}
\textit{Fault impedance, noise, network scale, and inference time.} Accuracy is reported in ten logarithmic $R_f$ bins between 0.001 and 100~$\Omega$ and at five noise levels from 0 to 2\%. Faults with $R_f>10~\Omega$ are counted as high-impedance faults (HIF). The identical protocol is applied to the three feeders and to DER-Fault, and per-sample inference time is measured for all classifiers.

\textit{Distribution shift.} Models are tested under load shift (trained on load scaling below 1.0, tested above), noise mismatch between training and test, the loss of one or two PMUs, and, on DER-Fault, training without DER (S1) and testing on S2--S6.
\end{nw}

\textit{Label-permutation test.} Class labels are randomly permuted prior to training; near-chance accuracy confirms that performance arises from genuine voltage patterns rather than data leakage.

\subsection{Evaluation metrics}\label{subsec4_3}

\begin{nw}
Accuracies are reported as mean $\pm$ standard deviation over the cross-validation folds wherever several folds are available. Performance is evaluated by overall classification accuracy and macro-averaged F1 score under stratified five-fold cross-validation, repeated with two random seeds for the placement study; HIF accuracy, and accuracy after the loss of one PMU (N-1) are also reported. Classifiers are compared with the Friedman test, and paired methods with the Wilcoxon signed-rank test on identical folds and noise realisations. The Bayes error is bounded from below by the Cover--Hart inequality applied to the leave-one-out nearest-neighbour error \cite{coverhart1967}. The purpose of the benchmarks is not to maximise absolute accuracy, but to characterise the behaviour and limits of snapshot-based diagnosis under controlled variations.
\end{nw}

\section{Results and discussion}\label{sec5}

\begin{nw}
This section answers the question posed in section~\ref{sec1}, namely where the error of snapshot-based diagnosis comes from and what removes it. Section~\ref{subsec5_0} separates the contributions of the classifier, the features and the fault resistance, and shows that the remaining error is caused by measurement error on high-impedance faults. Section~\ref{subsec5_err} quantifies this effect with the framework of section~\ref{subsec3_err} and gives the measurement-device accuracy class that a target accuracy requires. Section~\ref{subsec5_6} confirms that the conclusions do not depend on the classifier or on the size of the feeder. Section~\ref{subsec5_1} compares DMP with the other placement methods, including a reproduction of the conditions of a recent study. Section~\ref{subsec5_gen} tests the results on the DER-Fault data set and under distribution shift, and extends the snapshot to faulted-zone identification and to a reject option. Section~\ref{subsec5_10} combines these results into an operating envelope for deployment.
\end{nw}

\begin{figure}[t]
\begin{nwf}
\centering
\includegraphics[width=\textwidth]{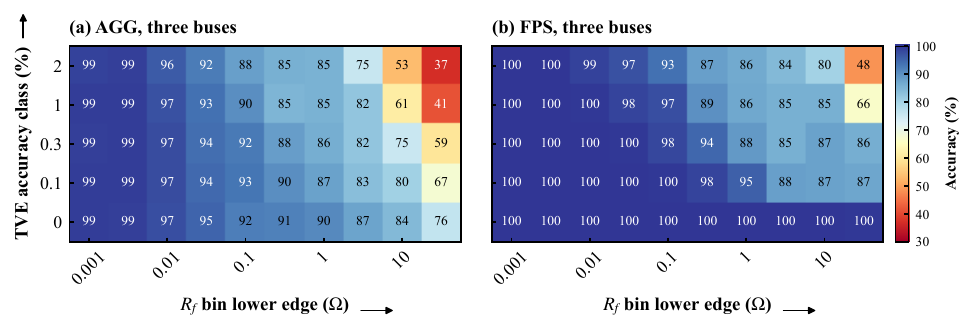}
\caption{Accuracy (\%) by fault resistance and TVE accuracy class, three buses, IEEE 13-node feeder. (a) AGG; (b) FPS.}\label{fig4}
\end{nwf}
\end{figure}

\begin{nw}
\subsection{Where the error comes from}\label{subsec5_0}

Table~\ref{tab_fs} compares the four feature sets on the 13-node feeder at 1\% TVE. Each cell gives the mean $\pm$ standard deviation over ten folds; bold marks the best classifier in each column, and the last row gives the Cover--Hart upper bound on the accuracy of any classifier. With all 14 buses monitored, LightGBM reaches 83.3\% with AGG, 90.6\% with PPA, 90.5\% with SEQ and 92.7\% with FPS; with the three-bus subset of figure~\ref{fig1}, the values are 83.4\%, 88.2\%, 88.5\% and 90.9\%. The fold-to-fold standard deviation is 0.5--2.2 points, so the gain of 7--9 points from AGG to FPS is several times larger than the spread, and it is significant (Wilcoxon $p=0.002$ over ten folds in both cases). For a given feature set, the three tree-based ensembles lie within one standard deviation of each other; the largest departures are LR on PPA and kNN on FPS with three buses.

Without noise, FPS classifies 99.7--100\% of the test samples correctly in every $R_f$ bin (figure~\ref{fig4}(b), bottom row), whereas AGG still reaches only 90.7\% with three buses because pooling over phases discards information. The errors of FPS under noise are therefore not a limitation of the classifiers: they arise when noise is comparable to the voltage deviation produced by the fault. The Cover--Hart estimate supports this. Its lower bound on the Bayes error is 5.5--10.6\% for the four feature sets at the 1\% TVE accuracy class, so no classifier can exceed roughly 89--94\% accuracy with these inputs, and the observed accuracies lie within a few points of that ceiling.

\end{nw}
\begin{table}[!htbp]
\centering
\begin{nwf}
\caption{Accuracy (\%) of seven classifiers with four feature sets, IEEE 13-node feeder, 1\% TVE accuracy class.}\label{tab_fs}
\footnotesize
\setlength{\tabcolsep}{3.2pt}
\begin{tabular}{lcccccccc}
\toprule
 & \multicolumn{4}{c}{All 14 buses monitored} & \multicolumn{4}{c}{Three monitored buses (figure~\ref{fig1})} \\
\cmidrule(lr){2-5}\cmidrule(lr){6-9}
Classifier & AGG & PPA & SEQ & FPS & AGG & PPA & SEQ & FPS \\
\midrule
RF       & \textbf{83.7}$\pm$1.2 & 90.5$\pm$0.7 & 90.4$\pm$0.6 & 92.6$\pm$0.7 & 83.5$\pm$0.9 & 88.3$\pm$0.8 & 88.5$\pm$0.6 & \textbf{91.2}$\pm$0.8 \\
XGBoost  & 83.6$\pm$0.9 & \textbf{90.6}$\pm$0.8 & \textbf{90.8}$\pm$0.6 & 92.6$\pm$0.6 & \textbf{83.5}$\pm$1.0 & \textbf{88.5}$\pm$0.8 & 88.5$\pm$0.6 & 90.6$\pm$0.7 \\
LightGBM & 83.3$\pm$1.1 & 90.6$\pm$0.8 & 90.5$\pm$0.8 & \textbf{92.7}$\pm$0.7 & 83.4$\pm$0.8 & 88.2$\pm$0.7 & \textbf{88.5}$\pm$0.5 & 90.9$\pm$0.7 \\
MLP      & 78.7$\pm$1.3 & 85.2$\pm$1.0 & 85.9$\pm$1.8 & 89.9$\pm$2.2 & 80.8$\pm$1.2 & 85.5$\pm$1.4 & 85.4$\pm$1.0 & 87.6$\pm$1.6 \\
SVM      & 79.2$\pm$1.3 & 80.2$\pm$1.1 & 85.6$\pm$1.2 & 85.4$\pm$1.1 & 81.0$\pm$1.0 & 80.8$\pm$1.1 & 85.1$\pm$1.1 & 83.3$\pm$0.9 \\
kNN      & 81.7$\pm$0.9 & 89.4$\pm$0.6 & 89.7$\pm$0.8 & 87.9$\pm$0.8 & 81.4$\pm$1.3 & 87.7$\pm$0.7 & 87.7$\pm$0.8 & 83.4$\pm$1.1 \\
LR       & 78.1$\pm$1.5 & 66.6$\pm$1.1 & 87.0$\pm$0.9 & 87.5$\pm$0.9 & 80.4$\pm$1.0 & 70.5$\pm$1.7 & 86.7$\pm$0.7 & 83.9$\pm$0.9 \\
\midrule
Cover--Hart bound & 89.4 & 94.4 & 94.5 & 93.7 & 89.7 & 93.2 & 93.1 & 91.1 \\
\bottomrule
\end{tabular}
\end{nwf}
\end{table}
\begin{nw}

Figure~\ref{fig4} resolves the accuracy by measurement error and fault resistance. Its columns are ten logarithmic $R_f$ bins labelled by their lower edge, and its rows are TVE accuracy classes (section~\ref{subsec3_err}). At the 1\% accuracy class, FPS classifies 98--100\% of the faults below 1~$\Omega$, and accuracy falls in the higher bins to 85\% between 10 and 32~$\Omega$ and to 66\% between 32 and 100~$\Omega$. FPS is more accurate than AGG in every bin, by up to 25 points in the 10--32~$\Omega$ bin (figure~\ref{fig4}(a, b)). The high-impedance limit is therefore a limit of measurement error, because a fault of several tens of ohms produces a voltage change of the same order as the error of a 1\% device. Lowering the error moves the limit upwards, as equation~(\ref{eq_rfstar}) predicts. At the 0.1\% accuracy class the two highest bins reach 87\%, against 85\% and 66\% at 1\%, and without error FPS reaches 99.7\% in the highest bin. Between the 0.1\% accuracy class and no error, the two highest bins stay near 87\%, which indicates that with three sensors part of these faults is limited by the sensor sites rather than by the measurement error.

\end{nw}

\subsubsection{Accuracy by fault resistance}\label{subsec5_7}

\begin{nw}
Figure~\ref{fig4} replaces the original table~2, which used AGG, two classifiers and $\pm$2\% noise. At the 1\% TVE accuracy class with FPS, faults below 10~$\Omega$ are classified with 85--100\% accuracy, while the 10--32~$\Omega$ and 32--100~$\Omega$ bins reach 85\% and 66\%. High-impedance faults thus produce voltage changes too small for reliable snapshot-based discrimination with a 1\% device, and this regime becomes more common as inverter-based resources limit fault currents. Over the three IEEE feeders with the reference three-bus sets and FPS, faults below 10~$\Omega$ are classified with 86--99\% accuracy at the 1\% TVE accuracy class, while faults of 10--100~$\Omega$ reach 46--92\% at the 1\% TVE accuracy class and 86--99\% with micro-PMU noise.
\end{nw}

\subsubsection{Feature ablation}\label{subsec5_2}

Removing each feature in turn leads to consistent but non-catastrophic changes \hlnew{(original figure 3(b))}. The largest degradations occur when removing $f_4 = \Vert\Delta V\Vert^{2}$ (RF: 87.0\% $\rightarrow$ 83.0\%) and $f_2$ (SVM: 85.0\% $\rightarrow$ 81.0\%), confirming that fault-induced voltage deviations are the primary source of class separability. \hlnew{Within the four statistics, therefore, the deviation terms carry most of the information; section~\ref{subsec5_0} shows that the larger loss comes from pooling over phases, which no choice among $f_1$--$f_4$ can recover.}

\subsubsection{Where misclassifications occur}\label{subsec5_5}

\begin{nw}
Misclassifications concentrate among faults with large $R_f$ (figure~\ref{fig4}), where the snapshot of every fault type approaches the pre-fault steady state. Such ambiguity cannot be resolved from a single snapshot without temporal or directional information \cite{lotfifard2011,karabulut2025}.
\end{nw}

\begin{nw}
\subsection{Measurement error and required measurement-device accuracy class}\label{subsec5_err}
Section~\ref{subsec5_0} located the remaining error in measurement error on high-impedance faults. This subsection measures how much accuracy each class of measurement device delivers and which class a target accuracy requires.
\end{nw}

\begin{nw}
\subsubsection{Two reference levels of measurement error}\label{subsec5_noise}
Table~\ref{tab5} compares the two noise levels. It uses LightGBM and gives the mean $\pm$ standard deviation over five folds. With micro-PMU noise, three sensors give 97.6\%, 99.3\% and 95.3\% with FPS on the three feeders, 3--9 points above the 1\% TVE accuracy class. The gain is largest for SEQ and FPS on the 13- and 123-node feeders, whose sequence and angle features are the most sensitive to measurement error.

\end{nw}
\begin{table}[!htbp]
\centering
\begin{nwf}
\caption{Accuracy (\%) with three PMUs at the 1\% TVE accuracy class and at measured micro-PMU noise.}\label{tab5}
\footnotesize
\begin{tabular}{lcccccc}
\toprule
 & \multicolumn{3}{c}{1\% TVE} & \multicolumn{3}{c}{$\mu$PMU (0.02\%, 0.002$^\circ$)} \\
\cmidrule(lr){2-4}\cmidrule(lr){5-7}
Network & AGG & SEQ & FPS & AGG & SEQ & FPS \\
\midrule
IEEE 13-node  & 83.4$\pm$1.0 & 88.4$\pm$0.6 & 91.0$\pm$0.6 & 89.1$\pm$1.1 & 95.3$\pm$0.9 & 97.6$\pm$0.7 \\
IEEE 34-node  & 93.0$\pm$0.4 & 95.8$\pm$0.4 & 96.3$\pm$0.3 & 93.9$\pm$0.6 & 97.8$\pm$0.4 & 99.3$\pm$0.1 \\
IEEE 123-node & 72.2$\pm$1.0 & 74.1$\pm$1.3 & 85.9$\pm$0.9 & 78.2$\pm$0.6 & 84.2$\pm$0.7 & 95.3$\pm$0.4 \\
\bottomrule
\end{tabular}
\end{nwf}
\end{table}

\subsubsection{Required measurement accuracy}\label{subsec5_spec}

\begin{nw}
Table~\ref{tab_spec} turns the detectability limit into a specification. For three DMP sensors, FPS and LightGBM, the TVE accuracy class was swept from 0.0025\% to 3\% and the table gives the largest class that still reaches each target accuracy. On the 34-node feeder, whose long lines produce large voltage changes, even the 3\% accuracy class gives 95\%. On the 13-node feeder, 90\% needs the 1\% accuracy class and 95\% needs 0.1\%, and on the 123-node feeder 95\% needs 0.02\%, which is the range of distribution micro-PMUs. A 99\% target is reached only with micro-PMU-grade devices and is not reached on the 123-node feeder within the tested range. The definition of the class matters as much as the class itself. With the common reading of 1\% TVE (table~\ref{tab_noise}), the same three sensors give 87.3\%, 95.4\% and 85.2\% on the three feeders, against 91.7\%, 96.5\% and 89.9\% for a device that actually meets the 1\% limit; the common reading behaves like the 3\% accuracy class.
\end{nw}

\begin{table}[!htbp]
\centering
\caption{\hlnew{Largest TVE accuracy class (\%) that reaches a target accuracy with three DMP sensors, and accuracy (\%, mean $\pm$ standard deviation over two seeds $\times$ five folds) at selected classes.}}\label{tab_spec}
\small
\begin{tabular}{lccc}
\toprule
Target accuracy & IEEE 13-node & IEEE 34-node & IEEE 123-node\\
\midrule
90\% & 1.0 & $\geq$3.0 & 0.5\\
95\% & 0.1 & $\geq$3.0 & 0.02\\
99\% & 0.005 & 0.02 & not reached\\
\midrule
Accuracy at the 1\% accuracy class & 91.7$\pm$0.9 & 96.5$\pm$0.4 & 89.9$\pm$0.6\\
Accuracy at the 3\% accuracy class & 87.0$\pm$1.0 & 95.3$\pm$0.3 & 84.6$\pm$0.7\\
Accuracy, common reading of 1\% TVE & 87.3$\pm$1.0 & 95.4$\pm$0.3 & 85.2$\pm$0.9\\
\bottomrule
\end{tabular}
\end{table}

\subsection{Classifiers and network scale}\label{subsec5_6}

\begin{nw}
Table~\ref{tab1} summarises all thirteen classifiers and the nearest-template classifier on the three feeders with FPS and three PMUs. The three PMUs are the hand-picked subset on the 13-node feeder and the topology rule on the 34- and 123-node feeders; values are the mean $\pm$ standard deviation over five folds, except for the nearest-template classifier, which has no training step and is evaluated once. Tree-based ensembles (RF, ExtraTrees, XGBoost, LightGBM, stacking) are the most accurate and lie within 1 point of each other in every setting; the Friedman test rejects equal performance across all classifiers ($p<10^{-30}$), and the mean ranks are 1.9 (ExtraTrees), 2.8 (stacking), 3.1 (RF), 3.7 (XGBoost) and 3.8 (LightGBM). The CNN1D does not outperform the tree-based ensembles, and linear or Gaussian models fail on the 123-node feeder (39--64\%) because the FPS features are not linearly separable there.

\end{nw}
\begin{table}[!htbp]
\centering
\begin{nwf}
\caption{Accuracy (\%) of all classifiers with FPS and three PMUs at two noise levels.}\label{tab1}
\small
\setlength{\tabcolsep}{3.5pt}
\begin{tabular}{llcccccc}
\toprule
 & & \multicolumn{2}{c}{IEEE 13-node} & \multicolumn{2}{c}{IEEE 34-node} & \multicolumn{2}{c}{IEEE 123-node} \\
\cmidrule(lr){3-4}\cmidrule(lr){5-6}\cmidrule(lr){7-8}
\hlnew{Family} & Classifier & 1\% TVE & $\mu$PMU & 1\% TVE & $\mu$PMU & 1\% TVE & $\mu$PMU \\
\midrule
\multirow{5}{*}{\hlnew{Tree ensemble}} & RF & 91.1$\pm$1.0 & 97.7$\pm$0.7 & 96.5$\pm$0.3 & 99.4$\pm$0.1 & 85.6$\pm$1.2 & 94.9$\pm$0.6 \\
 & ExtraTrees & 91.4$\pm$0.9 & 97.6$\pm$0.7 & 96.7$\pm$0.3 & 99.5$\pm$0.1 & 85.9$\pm$0.7 & 95.4$\pm$0.5 \\
 & XGBoost & 90.7$\pm$0.8 & 97.7$\pm$0.5 & 96.3$\pm$0.5 & 99.3$\pm$0.1 & 86.0$\pm$0.8 & 95.2$\pm$0.2 \\
 & LightGBM & 91.0$\pm$0.6 & 97.6$\pm$0.6 & 96.3$\pm$0.3 & 99.3$\pm$0.1 & 85.9$\pm$0.8 & 95.3$\pm$0.3 \\
 & Stacking & 91.2$\pm$1.0 & 97.6$\pm$0.6 & 96.4$\pm$0.2 & 99.4$\pm$0.1 & 86.4$\pm$0.9 & 95.4$\pm$0.4 \\
\cmidrule(l){1-8}
\hlnew{Kernel} & SVM & 83.2$\pm$1.1 & 84.2$\pm$1.4 & 95.1$\pm$0.3 & 95.2$\pm$0.3 & 74.3$\pm$0.8 & 78.7$\pm$0.8 \\
\cmidrule(l){1-8}
\multirow{2}{*}{\hlnew{Linear}} & LR & 83.8$\pm$1.1 & 85.9$\pm$1.5 & 96.1$\pm$0.4 & 96.8$\pm$0.3 & 52.1$\pm$1.0 & 58.0$\pm$1.5 \\
 & LDA & 79.1$\pm$1.6 & 78.8$\pm$2.0 & 93.8$\pm$0.5 & 93.8$\pm$0.5 & 51.9$\pm$1.0 & 53.1$\pm$1.2 \\
\cmidrule(l){1-8}
\hlnew{Instance-based} & kNN & 83.1$\pm$1.3 & 94.3$\pm$1.1 & 96.0$\pm$0.4 & 98.6$\pm$0.1 & 68.0$\pm$1.1 & 88.0$\pm$0.7 \\
\cmidrule(l){1-8}
\multirow{2}{*}{\hlnew{Probabilistic}} & QDA & 73.5$\pm$1.5 & 73.6$\pm$1.5 & 90.4$\pm$0.2 & 90.4$\pm$0.2 & 63.7$\pm$0.8 & 57.4$\pm$1.1 \\
 & NB & 67.5$\pm$2.0 & 76.5$\pm$1.8 & 94.3$\pm$0.2 & 96.1$\pm$0.2 & 42.8$\pm$1.2 & 39.1$\pm$0.5 \\
\cmidrule(l){1-8}
\multirow{2}{*}{\hlnew{Neural network}} & MLP & 88.7$\pm$1.0 & 89.9$\pm$3.1 & 95.7$\pm$0.3 & 97.4$\pm$0.1 & 83.9$\pm$0.5 & 92.6$\pm$0.6 \\
 & CNN1D & 83.8$\pm$1.6 & 85.3$\pm$1.4 & 95.2$\pm$0.7 & 96.3$\pm$0.2 & 74.8$\pm$2.4 & 81.2$\pm$0.9 \\
\cmidrule(l){1-8}
\hlnew{No training} & Nearest template & 86.0 & 89.1 & 94.6 & 96.1 & 75.5 & 84.1 \\
\bottomrule
\end{tabular}
\end{nwf}
\end{table}
\begin{nw}

Accuracy does not decrease monotonically with feeder size. The 34-node feeder is the easiest (96\% at the 1\% TVE accuracy class) because its long lines produce large, well-separated voltage deviations, while the 123-node feeder is the hardest. On the 123-node feeder the three topology-chosen sensors reach 85.9\% with FPS, against 89.8\% for the three DMP sensors (table~\ref{tab4}), so the information is present in the snapshot but depends on where it is measured.
\end{nw}

\subsubsection{Cross-seed robustness and label-permutation test}\label{subsec5_3}

\begin{nw}
Across ten folds, the standard deviation of accuracy of the tree-based ensembles is 0.8--1.5 points for every feature set (table~\ref{tab_fs}). When labels are shuffled, accuracy collapses to $33.1\pm0.8\%$, $34.2\pm1.8\%$ and $33.4\pm2.6\%$ on the 13-, 34- and 123-node feeders (table~\ref{tab6}), i.e.\ chance level for three classes, confirming that the reported performance is not an artefact of leakage.
\end{nw}

\subsection{Sensor placement}\label{subsec5_1}

\begin{figure}[t]
\begin{nwf}
\centering
\includegraphics[width=0.85\textwidth]{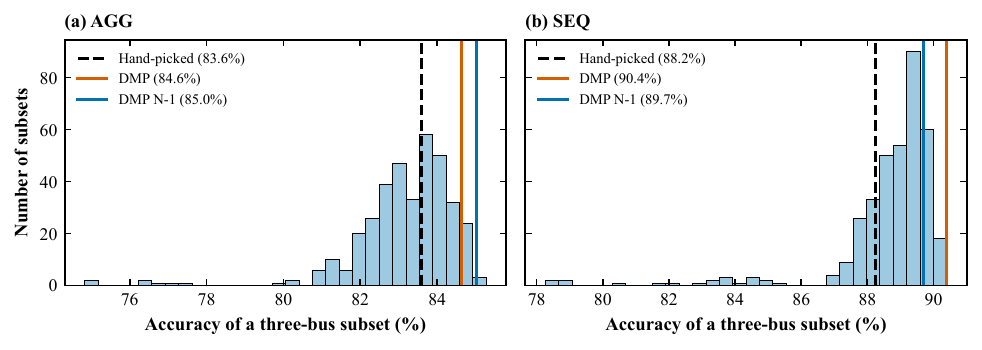}
\caption{Accuracy of all 364 three-bus subsets of the IEEE 13-node feeder at the 1\% TVE accuracy class. (a) AGG; (b) SEQ.}\label{fig5}
\end{nwf}
\end{figure}

\begin{nw}
On the 13-node feeder, the three-bus subset of figure~\ref{fig1} loses 2--4 points against all 14 buses (86.3\% against 88.2\% with FPS, table~\ref{tab_fs}). Snapshot voltage measurements are therefore highly redundant. Placement still matters. Over all 364 three-bus subsets at the 1\% TVE accuracy class, accuracy ranges from 74.8\% to 85.3\% with AGG and from 78.3\% to 90.4\% with SEQ (figure~\ref{fig5}). The hand-picked subset ranks 152nd with AGG and 287th with SEQ, while the DMP subset (634, 680, 692) ranks 23rd with AGG and first with SEQ.

\end{nw}
\begin{nw}

Table~\ref{tab4} compares DMP with the placement baselines and with the distinguishability-based placement of \cite{kaliuzhnyi2027statistical} for three sensors and FPS. The classifier is LightGBM at the corrected 1\% TVE accuracy class, accuracy is the mean $\pm$ standard deviation over two seeds $\times$ five folds, and HIF is the accuracy on faults with $R_f>10~\Omega$. The method of \cite{kaliuzhnyi2027statistical} is applied to the same voltage-only snapshot, with its whitened-phasor distance over all pairs of fault scenarios of different type and the fault-resistance range of the original study ($R_f\le10~\Omega$). The well-placed methods lie within about one point of each other in overall accuracy on every feeder (91.0--91.9\%, 96.3--96.6\% and 89.0--90.1\%), so at this noise level placement mainly avoids poor sites rather than finding a uniquely good one. The differences are larger on high-impedance faults. On the 13-node feeder DMP gives the highest HIF accuracy (79.7\%), against 75.6--78.8\% for the other model-based methods and 76.9\% for the distinguishability-based placement. On the 34-node feeder all methods give 91.0--92.1\%. On the 123-node feeder DMP (77.1\%) is close to the best baselines (MRA ILP 78.3\%, MI greedy 78.0\%) and above the distinguishability-based placement (73.9\%) and FSNR-SVM (75.1\%). Losing one of the three DMP sensors lowers the accuracy by at most 0.6 points (to 91.5\%, 96.6\% and 89.1\% in the worst case), and the N-1-constrained design gives the same accuracy as plain DMP, because the three DMP sites already overlap in what they observe. DMP needs no trained classifier and selects sensors in under one second, against 1.9--120~s for FSNR-SVM and GA-LDA, which train a classifier for every candidate set. Paired Wilcoxon tests over the ten folds against the margin-based design confirm this pattern. On the 13- and 123-node feeders the topology rule, the observability programme and FSNR-SVM are significantly less accurate ($p\le0.004$), whereas MI greedy and SA coverage are not significantly different ($p\ge0.08$); on the 34-node feeder only the observability programme and the topology rule differ ($p=0.02$ and 0.045).

\end{nw}
\begin{table}[!htbp]
\centering
\begin{nwf}
\caption{\hlnew{Placement of three PMUs at the corrected 1\% TVE accuracy class: accuracy and HIF accuracy (\%).}}\label{tab4}
\small
\begin{tabular}{lcccccc}
\toprule
 & \multicolumn{2}{c}{IEEE 13-node} & \multicolumn{2}{c}{IEEE 34-node} & \multicolumn{2}{c}{IEEE 123-node} \\
\cmidrule(lr){2-3}\cmidrule(lr){4-5}\cmidrule(lr){6-7}
Method & Acc. & HIF & Acc. & HIF & Acc. & HIF \\
\midrule
DMP (proposed) & 91.9$\pm$0.8 & 79.7 & 96.4$\pm$0.4 & 91.4 & 89.8$\pm$0.7 & 77.1 \\
DMP N-1 (proposed) & 91.6$\pm$0.7 & 79.9 & 96.3$\pm$0.4 & 91.0 & 89.8$\pm$0.7 & 77.2 \\
MRA ILP & 91.2$\pm$0.8 & 75.6 & 96.4$\pm$0.3 & 91.2 & 90.1$\pm$0.7 & 78.3 \\
QR pivot & 91.5$\pm$0.8 & 78.5 & 96.3$\pm$0.4 & 91.2 & 89.5$\pm$0.6 & 76.7 \\
D-optimal & 91.5$\pm$0.7 & 78.8 & 96.6$\pm$0.3 & 92.0 & 89.6$\pm$0.8 & 76.7 \\
MI greedy & 91.8$\pm$0.6 & 79.4 & 96.4$\pm$0.3 & 91.4 & 90.1$\pm$0.7 & 78.0 \\
FSNR-SVM & 91.1$\pm$0.7 & 75.7 & 96.5$\pm$0.2 & 91.7 & 89.3$\pm$0.9 & 75.1 \\
GA-LDA & 91.5$\pm$0.6 & 76.9 & 96.5$\pm$0.2 & 91.0 & 89.9$\pm$0.6 & 77.8 \\
SA coverage & 91.7$\pm$0.8 & 78.6 & 96.6$\pm$0.3 & 92.1 & 90.0$\pm$0.6 & 77.8 \\
\hlnew{Max--min Mahalanobis \cite{kaliuzhnyi2027statistical}} & 91.0$\pm$1.0 & 76.9 & 96.3$\pm$0.3 & 91.4 & 89.0$\pm$0.9 & 73.9 \\
\bottomrule
\end{tabular}
\end{nwf}
\end{table}
\begin{figure}[t]
\begin{nwf}
\centering
\includegraphics[width=\textwidth]{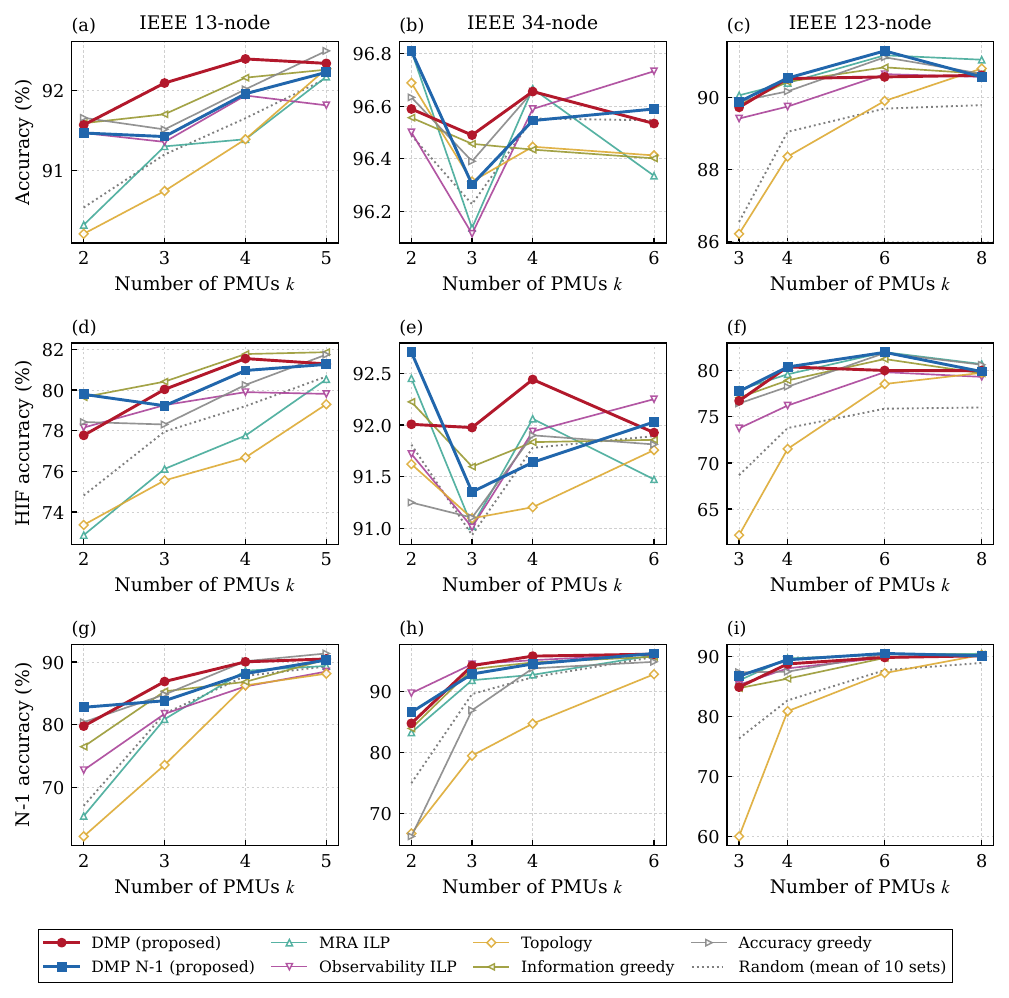}
\caption{Accuracy (a)--(c), HIF accuracy (d)--(f) and N-1 accuracy (g)--(i) against the number of PMUs.}\label{fig6}
\end{nwf}
\end{figure}
\begin{nw}

The differences among the well-placed methods are small compared with the effect of measurement error and features, which are examined in the following subsections.

\textit{Ranking candidate sets before installation.} On 40 random sensor sets per feeder, the index $B(S)$ of equation~(\ref{eq10}) is rank-correlated with the measured error (Spearman rank correlation $r_s=0.94$ on the 13-node feeder, 0.93 on the 123-node feeder and 0.61 on the 34-node feeder, $p<10^{-4}$ in every case). The measured error exceeded $B(S)$ for every set, so $B(S)$ ranks candidates but does not bound the error. Minimising $B(S)$ directly as a placement objective did not improve on DMP.
\end{nw}

\subsubsection{Comparison under the conditions of a previous study}\label{subsec5_repro}

\begin{nw}
To compare with \cite{khattak2025} on equal terms, its conditions were reproduced on the same feeders and sensor counts, namely four sensors on the 34-node feeder and nine on the 123-node feeder, fault resistances of 0.01--10~$\Omega$ and no measurement noise. The sensor sets of the FSNR-SVM method of \cite{khattak2025}, of DMP and of five random draws were evaluated with the same FPS features and LightGBM (table~\ref{tab_repro}). Under these conditions every sensor set classifies 99.9--100\% of the faults, so they cannot distinguish a good placement from a random one. The same holds with micro-PMU noise over the full range of fault resistance. With four or nine sensors the three sets also remain within one standard deviation of each other at the 1\% TVE accuracy class, so the benefit of placement appears mainly with fewer sensors, as in table~\ref{tab4}, and on high-impedance faults.
\end{nw}

\begin{table}[!htbp]
\centering
\caption{\hlnew{Accuracy (\%, mean $\pm$ standard deviation over two seeds $\times$ five folds) under the conditions of \cite{khattak2025} and under measurement error, FPS and LightGBM.}}\label{tab_repro}
\small
\begin{tabular}{lcccccc}
\toprule
 & \multicolumn{3}{c}{IEEE 34-node, 4 sensors} & \multicolumn{3}{c}{IEEE 123-node, 9 sensors}\\
\cmidrule(lr){2-4}\cmidrule(lr){5-7}
Sensor set & No noise, & $\mu$PMU, & 1\% TVE, & No noise, & $\mu$PMU, & 1\% TVE,\\
 & $R_f\le10~\Omega$ & all $R_f$ & all $R_f$ & $R_f\le10~\Omega$ & all $R_f$ & all $R_f$\\
\midrule
FSNR-SVM \cite{khattak2025} & 99.97$\pm$0.06 & 99.27$\pm$0.09 & 96.68$\pm$0.24 & 99.98$\pm$0.04 & 95.79$\pm$0.49 & 90.82$\pm$0.83\\
DMP & 99.94$\pm$0.09 & 99.37$\pm$0.12 & 96.63$\pm$0.35 & 99.92$\pm$0.09 & 95.73$\pm$0.36 & 91.06$\pm$0.73\\
Random (5 sets) & 99.96$\pm$0.06 & 99.35$\pm$0.17 & 96.65$\pm$0.28 & 99.93$\pm$0.08 & 95.80$\pm$0.41 & 90.41$\pm$0.63\\
\bottomrule
\end{tabular}
\end{table}

\subsubsection{Selection and inference time}\label{subsec5_9}

\begin{nw}
All classifiers complete single-sample inference in under 25~ms on a cloud central processing unit (CPU) with four virtual cores: MLP 643~$\mu$s, XGBoost 669~$\mu$s, SVM 671~$\mu$s, LR 685~$\mu$s, kNN 1.3~ms, LightGBM 1.4~ms and RF 23~ms. In batch mode, the cost falls to 0.4--154~$\mu$s per sample. DMP selects sensors in under 1~s offline. These times support use as fast first-stage screening tools on resource-constrained hardware. Table~\ref{tab_cost} compares the time each placement method needs to select three sensors, after the design fault scenarios have been simulated. DMP and the other model-based methods need less than two seconds on every feeder, whereas the methods that train a classifier for each candidate set (FSNR-SVM and GA-LDA) or estimate mutual information need several seconds, and FSNR-SVM needs 43~s on the 123-node feeder and 133~s for nine sensors.
\end{nw}

\begin{table}[!htbp]
\centering
\caption{\hlnew{Time (s) to select three sensors, excluding the simulation of the design fault scenarios.}}\label{tab_cost}
\small
\begin{tabular}{lcccc}
\toprule
Method & Trained classifier & IEEE 13-node & IEEE 34-node & IEEE 123-node\\
\midrule
DMP (proposed) & no & 0.01 & 0.02 & 0.19\\
DMP N-1 (proposed) & no & 0.01 & 0.03 & 1.66\\
MRA ILP & no & 0.01 & 0.04 & 0.17\\
QR pivot & no & $<$0.01 & 0.01 & 0.13\\
D-optimal & no & 0.01 & 0.03 & 0.32\\
SA coverage & no & 0.14 & 0.15 & 0.24\\
MI greedy & no & 1.13 & 2.81 & 10.23\\
GA-LDA & yes & 3.29 & 6.10 & 8.57\\
FSNR-SVM & yes & 4.66 & 5.95 & 43.12\\
\bottomrule
\end{tabular}
\end{table}
\begin{nw}
\end{nw}

\begin{nw}
\subsection{Generalisation and extended outputs}\label{subsec5_gen}
The results so far were obtained on simulated feeders with a fixed operating point and for the fault type only. This subsection tests them on an independent data set and under distribution shift, and then asks what else the same snapshot can deliver, namely the faulted zone and an estimate of when the diagnosis should not be trusted.
\end{nw}

\begin{nw}
\subsubsection{Validation on the DER-Fault data set}\label{subsec5_7b}

Table~\ref{tab_der} reports these results. Its left part gives seven classifiers trained and tested on all DER scenarios S1--S6 (mean $\pm$ standard deviation over five folds; bold: best classifier per feature set). Its right part gives the range of LightGBM accuracy over the DER scenarios when trained on all scenarios, and when trained on S1 (no DER) and tested on S2--S6. On DER-Fault with all 93 PMU-capable buses and 1\% TVE, LightGBM reaches 87.3\% (AGG), 92.0\% (PPA), 91.3\% (SEQ) and 88.4\% (FPS); without noise the values are 94.0\%, 99.8\%, 99.6\% and 99.9\%. The ordering differs from the IEEE feeders: with 93 buses, FPS has several hundred features, and at 1\% noise the per-phase aggregates of PPA average out the noise better than the per-node features of FPS. A model trained only on S1 (no DER) and tested on S2--S6 keeps 89--90\% with SEQ, 86--87\% with FPS and 81--84\% with AGG (table~\ref{tab_der}), so the sequence features transfer best to unseen DER penetrations. For sensor placement, DMP with 20 and 36 PMUs gives 88.5\% and 88.2\%, against 87.8\% and 88.0\% for the NSGA-II placements published with the data set; all placement methods with ten or more PMUs lie within about 1 point.

\end{nw}
\begin{table}[!htbp]
\centering
\begin{nwf}
\caption{Accuracy (\%) on the DER-Fault data set (Iowa 240-bus system, 1\% TVE accuracy class).}\label{tab_der}
\footnotesize
\setlength{\tabcolsep}{2.6pt}
\begin{tabular}{lccccccccc}
\toprule
 & \multicolumn{7}{c}{Trained and tested on S1--S6} & \multicolumn{2}{c}{LightGBM, by scenario} \\
\cmidrule(lr){2-8}\cmidrule(lr){9-10}
Set & RF & XGBoost & LightGBM & MLP & SVM & kNN & LR & S1--S6 & S1 $\rightarrow$ S2--S6 \\
\midrule
AGG & \textbf{87.7}$\pm$0.2 & 87.2$\pm$0.6 & 87.3$\pm$0.4 & 81.0$\pm$1.1 & 79.6$\pm$0.3 & 87.5$\pm$0.4 & 58.6$\pm$1.0 & 86.3--87.8 & 80.6--83.7 \\
PPA & \textbf{92.8}$\pm$0.1 & 92.0$\pm$0.3 & 92.0$\pm$0.2 & 90.2$\pm$0.7 & 84.5$\pm$0.2 & 91.9$\pm$0.2 & 69.1$\pm$0.4 & -- & -- \\
SEQ & \textbf{91.9}$\pm$0.2 & 91.3$\pm$0.2 & 91.3$\pm$0.2 & 89.7$\pm$0.8 & 88.1$\pm$0.6 & 90.9$\pm$0.3 & 87.8$\pm$0.6 & 90.4--91.9 & 89.4--90.1 \\
FPS & 87.9$\pm$0.4 & 88.1$\pm$0.4 & \textbf{88.4}$\pm$0.2 & 87.8$\pm$0.4 & 87.4$\pm$0.5 & 87.5$\pm$0.6 & 82.3$\pm$0.6 & 88.0--89.3 & 86.2--87.3 \\
\bottomrule
\end{tabular}
\end{nwf}
\end{table}

\begin{nw}

\subsubsection{Robustness to distribution shift and harder tasks}\label{subsec5_8}

With three DMP PMUs and FPS (table~\ref{tab6}), each entry is the mean $\pm$ standard deviation over five repetitions with different 70/30 train--test splits and noise realisations, with test noise of the 1\% TVE accuracy class; training at low load and testing at high load costs 1.2--2.8 points. Training with micro-PMU noise and testing at the 1\% TVE accuracy class costs about 4 points on the 13- and 123-node feeders against training at the 1\% TVE accuracy class; models should therefore be trained at the highest noise level expected in service. Losing one PMU costs 3--8 points; losing two of three reduces accuracy to 68--80\% on average, and the worst pair reaches only 55--67\%.

Under conditions comparable to \cite{khattak2025} (no noise, $R_f\le10~\Omega$, whereas \cite{khattak2025} uses bolted faults), four PMUs reach 99.9--100\% on the 34-node feeder whether they are placed by FSNR-SVM, by DMP or at random, which confirms that their setting does not discriminate between placements. With micro-PMU noise and the full $R_f$ range, the four-class task (adding double-line-to-ground (LLG) faults) and the ten-class phase-resolved task reach 97.0\% and 97.3\% on the 13-node feeder, 98.9\% and 99.2\% on the 34-node feeder, and 94.4\% and 94.2\% on the 123-node feeder with three PMUs and FPS.

\end{nw}
\begin{table}[!htbp]
\centering
\begin{nwf}
\caption{Robustness of three DMP PMUs with FPS and LightGBM (accuracy, \%).}\label{tab6}
\small
\begin{tabular}{lccc}
\toprule
Condition & 13-node & 34-node & 123-node \\
\midrule
Reference                           & 92.0$\pm$0.4 & 96.5$\pm$0.5 & 90.0$\pm$0.4 \\
Load shift (low $\rightarrow$ high) & 89.7$\pm$0.2 & 95.3$\pm$0.2 & 87.2$\pm$0.4 \\
Trained $\mu$PMU, tested 1\% TVE    & 88.0$\pm$0.2 & 95.8$\pm$0.6 & 86.2$\pm$0.4 \\
N-1 (mean over lost PMU)            & 86.3$\pm$2.0 & 93.5$\pm$2.2 & 81.7$\pm$0.5 \\
N-2 (mean over lost pairs)          & 70.7$\pm$2.5 & 80.0$\pm$5.3 & 67.7$\pm$1.1 \\
N-2 (worst pair)                    & 62.2$\pm$6.3 & 66.8$\pm$5.3 & 54.6$\pm$3.5 \\
Shuffled labels                     & 33.1$\pm$0.8 & 34.2$\pm$1.8 & 33.4$\pm$2.6 \\
\bottomrule
\end{tabular}
\end{nwf}
\end{table}

\subsubsection{Faulted-zone identification}\label{subsec5_zone}

\begin{nw}
The same snapshot also carries coarse location information. Each feeder was divided into zones of similar size by graph distance from the substation, and the classifier was trained on the zone label with the same three DMP sensors and features. Table~\ref{tab_zone} shows that two zones are identified with 91--96\% accuracy at the 1\% TVE accuracy class and three zones with about 85\%, against chance levels of 50\% and 33\%. Accuracy falls as the zones become finer, and with five zones it is 75--77\%. The snapshot of a few voltage sensors is therefore suited to locating the faulted part of the feeder, while the faulted line or the distance requires more sensors or other methods.
\end{nw}

\begin{table}[!htbp]
\centering
\caption{\hlnew{Faulted-zone accuracy (\%, mean $\pm$ standard deviation over two seeds $\times$ five folds) with three DMP sensors, at the 1\% TVE accuracy class and with micro-PMU noise.}}\label{tab_zone}
\small
\setlength{\tabcolsep}{3pt}
\begin{tabular}{lccccccc}
\toprule
 & & \multicolumn{2}{c}{IEEE 13-node} & \multicolumn{2}{c}{IEEE 34-node} & \multicolumn{2}{c}{IEEE 123-node}\\
\cmidrule(lr){3-4}\cmidrule(lr){5-6}\cmidrule(lr){7-8}
Zones & Chance & 1\% TVE & $\mu$PMU & 1\% TVE & $\mu$PMU & 1\% TVE & $\mu$PMU\\
\midrule
2 & 50.0 & 90.7$\pm$0.6 & 97.7$\pm$0.3 & 96.0$\pm$0.4 & 99.9$\pm$0.1 & 91.0$\pm$0.6 & 94.6$\pm$0.3\\
3 & 33.3 & 84.1$\pm$0.5 & 90.4$\pm$0.5 & 84.8$\pm$0.6 & 94.6$\pm$0.3 & 84.6$\pm$0.9 & 89.8$\pm$0.8\\
4 & 25.0 & 79.3$\pm$0.7 & 87.5$\pm$0.6 & 80.8$\pm$0.5 & 93.8$\pm$0.3 & 77.7$\pm$0.6 & 83.1$\pm$0.8\\
5 & 20.0 & 75.6$\pm$0.6 & 85.4$\pm$0.4 & 76.5$\pm$0.9 & 88.8$\pm$0.6 & 74.7$\pm$0.9 & 80.5$\pm$0.6\\
\bottomrule
\end{tabular}
\end{table}

\subsubsection{Reject option}\label{subsec5_reject}

\begin{nw}
A screening tool is more useful if it can recognise the snapshots it is likely to misclassify and pass them to a slower method. Two scores were used to rank the test snapshots, the largest class probability of LightGBM and a classifier-free margin, defined as the difference between the distances of the snapshot to the nearest training snapshots of the second closest and the closest fault type after the features are scaled by the measurement-noise standard deviation. Table~\ref{tab_reject} reports the share of snapshots that can be classified when the accuracy on them must reach 99\%, the share of all errors found among the 10\% least certain snapshots, and the share of high-impedance faults among those rejected snapshots. Deferring 13--16\% of the snapshots lifts the accuracy from 90--92\% to 99\% on the 13- and 123-node feeders, and 4\% suffices on the 34-node feeder. The rejected snapshots are 2.1--2.7 times richer in high-impedance faults than the whole data set, in line with the error decomposition of section~\ref{subsec5_0}. The classifier probability is slightly better than the margin, by about one point of coverage, while the margin needs neither a trained classifier nor calibrated probabilities.
\end{nw}

\begin{table}[!htbp]
\centering
\caption{\hlnew{Reject option with three DMP sensors at the 1\% TVE accuracy class (\%, mean $\pm$ standard deviation over two seeds $\times$ five folds). P: classifier probability; M: classifier-free margin.}}\label{tab_reject}
\small
\begin{tabular}{lcccccc}
\toprule
 & \multicolumn{2}{c}{IEEE 13-node} & \multicolumn{2}{c}{IEEE 34-node} & \multicolumn{2}{c}{IEEE 123-node}\\
\cmidrule(lr){2-3}\cmidrule(lr){4-5}\cmidrule(lr){6-7}
 & P & M & P & M & P & M\\
\midrule
Accuracy without rejection & \multicolumn{2}{c}{91.9$\pm$0.9} & \multicolumn{2}{c}{96.5$\pm$0.4} & \multicolumn{2}{c}{89.8$\pm$0.7}\\
Share classified at 99\% accuracy & 87.1$\pm$1.2 & 86.3$\pm$1.2 & 96.0$\pm$0.6 & 95.8$\pm$0.5 & 84.1$\pm$1.2 & 83.2$\pm$0.9\\
Errors among the 10\% rejected & 73.8$\pm$5.5 & 69.1$\pm$5.2 & 100$\pm$0.0 & 100$\pm$0.0 & 61.2$\pm$4.5 & 56.0$\pm$5.0\\
HIF among the 10\% rejected & 54.2$\pm$3.2 & 54.6$\pm$3.4 & 43.0$\pm$3.8 & 44.5$\pm$4.0 & 44.6$\pm$4.9 & 47.6$\pm$3.6\\
HIF in the whole data set & \multicolumn{2}{c}{20.1} & \multicolumn{2}{c}{20.3} & \multicolumn{2}{c}{20.4}\\
\bottomrule
\end{tabular}
\end{table}

\subsection{Discussion: an operating envelope for reliable deployment}\label{subsec5_10}

\begin{nw}
The results reveal four properties of snapshot-based fault diagnosis. First, the classifiers are not the bottleneck: without noise the full phasor snapshot is almost perfectly separable, and with noise the tree-based ensembles agree within 1 point and approach the Cover--Hart ceiling. Second, the original aggregate features lose 3--14 points because they pool over phases and discard angles; per-phase and sequence quantities recover most of this loss. Third, the accuracy limit is set jointly by measurement error and fault resistance: high-impedance faults fail at the 1\% TVE accuracy class but are largely recovered with micro-PMUs. The 3--9 point gap between the two device classes is itself a finding: studies that evaluate without measurement noise, such as \cite{khattak2025}, implicitly assume the micro-PMU end of this range, and the choice of measurement-device accuracy class therefore matters as much as the choice of classifier or placement. Fourth, a few well-placed sensors capture most of the information, and placement matters mainly to avoid poor sites and on high-impedance faults.

\end{nw}

Sequence-aware (e.g.\ convolutional or long short-term memory networks) and topology-aware (e.g.\ graph neural networks) models recover temporal or structural information that a single snapshot discards, at higher data and computational cost \cite{siddique2024,karabulut2025}. Rather than competing with them on accuracy, this work quantifies the intrinsic ceiling of the lightweight snapshot regime. Table~\ref{tab3} condenses the findings into an operating envelope that maps network scale and fault conditions to the regimes where snapshot-based diagnosis can be relied upon. It refers to three PMUs with FPS and LightGBM on the three IEEE feeders unless stated otherwise. The suitability of each condition is assigned from the lowest accuracy observed under it: reliable ($\ge$95\%), screening only (75--95\%) and not reliable ($<$75\%). Snapshot methods are best positioned as first-stage screening tools for small-to-medium feeders under low-to-moderate impedance, with escalation to time-domain or protection-grade analysis outside this envelope.

\begin{table}[!htbp]
\centering
\begin{nwf}
\caption{Operating envelope and deployment recommendations for snapshot-based fault diagnosis.}\label{tab3}
\footnotesize
\setlength{\tabcolsep}{4pt}
\begin{tabular}{@{}>{\raggedright\arraybackslash}p{0.34\textwidth}
                >{\raggedright\arraybackslash}p{0.13\textwidth}
                >{\raggedright\arraybackslash}p{0.47\textwidth}@{}}
\toprule
Condition & Suitability & Recommendation \\
\midrule
\multicolumn{3}{@{}l}{\textit{Measurement noise}} \\
Micro-PMU noise (0.02\%, 0.002$^\circ$) & Reliable & Preferred measurement-device accuracy class for snapshot diagnosis \\
1\% TVE, 34-node feeder & Reliable & Usable without further measures \\
1\% TVE, 13- and 123-node feeders & Screening only & Confirm by protection or time-domain analysis \\
Service noise above training noise & Screening only & Train at the highest noise level expected in service \\
\addlinespace
\multicolumn{3}{@{}l}{\textit{Fault resistance}} \\
$R_f<10~\Omega$, 1\% TVE & Screening only & Snapshot classification adequate for first-stage use \\
$R_f$ 10--100~$\Omega$, 1\% TVE & Not reliable & Escalate to time-domain analysis or use micro-PMUs \\
$R_f$ 10--100~$\Omega$, micro-PMU & Screening only & Acceptable for screening high-impedance faults \\
\addlinespace
\multicolumn{3}{@{}l}{\textit{Features and placement}} \\
Aggregate features only ($f_1$--$f_4$) & Not reliable & Use per-phase or phasor features instead \\
Topology-based placement, 123-node & Screening only & Place sensors by DMP before installation \\
\addlinespace
\multicolumn{3}{@{}l}{\textit{Sensor loss}} \\
One of three PMUs lost & Screening only & Use the DMP N-1 placement \\
Two of three PMUs lost & Not reliable & Install redundant PMUs; flag low-confidence outputs \\
\addlinespace
\multicolumn{3}{@{}l}{\textit{Operating conditions}} \\
Load level outside training range & Screening only & No retraining needed within the tested load range \\
DER level not seen in training & Screening only & Use SEQ features when the DER level may change \\
\bottomrule
\end{tabular}
\end{nwf}
\end{table}

\section{Conclusions}\label{sec6}

\begin{nw}
This paper studied where the error of snapshot-based fault diagnosis comes from and what removes it, on the IEEE 13-, 34- and 123-node feeders and the DER-Fault data set, with four feature sets, thirteen classifiers and twelve placement methods. Four findings stand out. First, measurement error must be stated as a TVE accuracy class defined by a percentile; the common reading of 1\% TVE as a standard deviation simulates a device close to the 3\% accuracy class and understates the achievable accuracy by 1--5 points. Second, the classifier is not the limiting factor, because the full phasor snapshot is almost perfectly separable without noise, and the remaining error comes from measurement error on high-impedance faults; three voltage sensors classify 90--96\% of faults at the 1\% accuracy class and 96--99\% with micro-PMUs, and the required measurement-device accuracy class follows from the target accuracy. Third, a discriminability-margin placement computed from simulations alone in under two seconds matches the best placement methods, selects the best of all 364 three-bus subsets on the 13-node feeder, and is more accurate on high-impedance faults than a recent distinguishability-based method on two of the three feeders. Fourth, a reject option that defers 4--16\% of the snapshots, mainly high-impedance faults, raises the accuracy on the rest to 99\%, and the same snapshot locates the faulted zone among three with about 85\% accuracy. Several limitations apply: the data are simulated with steady-state phasor models, so snapshot timing and transients are not represented; only voltage measurements are used; the micro-PMU noise level is taken from one measurement campaign; and the operating envelope holds under load shift and DER penetration but not under the loss of two of three sensors. Future work will add feeder-head current measurements, short temporal windows and validation on field recordings.
\end{nw}

\section*{Acknowledgments}
Not applicable.

\section*{Funding}
The authors received no specific funding for this work.

\section*{Data availability}
The code, the simulated data sets and the result files needed to reproduce all results and figures of this paper are available at \url{https://github.com/phungtu682/Snapshot-fault-diagnosis}. The DER-Fault data set is available on Zenodo (\url{https://doi.org/10.5281/zenodo.22904220}).

\section*{Conflict of interest}
The authors declare that they have no competing interests.

\end{document}